\documentclass[10pt,journal,compsoc]{IEEEtran}
\ifCLASSOPTIONcompsoc
  \usepackage[nocompress]{cite}
\else
  \usepackage{cite}
\fi
\ifCLASSINFOpdf
\else
\fi
\usepackage{amsmath}
\usepackage{float}
\restylefloat{table}
\usepackage[pdftex]{graphicx}
\usepackage{algorithm}
\usepackage{algpseudocode}
\usepackage{hyperref}

\DeclareMathOperator*{\argmax}{arg\,max} 
\begin{document}
%
\title{Self Paced Deep Learning for Weakly Supervised Object Detection}
%
%
%
%

\author{Enver~Sangineto$^\dagger$,
        Moin~Nabi$^\dagger$,
        Dubravko~Culibrk  
        and~Nicu~Sebe,
\IEEEcompsocitemizethanks{\IEEEcompsocthanksitem Enver Sangineto, Moin Nabi and Nicu Sebe are with the Department of Information Engineering and Computer Science (DISI),
University of Trento, Italy.\protect\\
E-mail: enver.sangineto@unitn.it; sebe@disi.unitn.it
\IEEEcompsocthanksitem Moin Nabi is also with SAP SE, Berlin, Germany.\protect\\
E-mail: m.nabi@sap.com 
\IEEEcompsocthanksitem Dubravko Culibrk is with the Department of Industrial Engineering and Management,
University of Novi Sad, Serbia.\protect\\
E-mail: dculibrk@uns.ac.rs}
\thanks{$\dagger$ These two authors contributed equally.}}

%
%

\markboth{Journal of \LaTeX\ Class Files,~Vol.~14, No.~8, August~2015}%
{Shell \MakeLowercase{\textit{et al.}}: Bare Demo of IEEEtran.cls for Computer Society Journals}
%



\IEEEtitleabstractindextext{%
\begin{abstract}
In a weakly-supervised scenario object detectors need to be trained using  image-level annotation
alone. 
Since bounding-box-level ground truth is not available, most of the solutions proposed so far are based on an iterative, 
 Multiple Instance Learning framework in which the 
current 
 classifier is used to select the highest-confidence boxes in each image, which are treated as pseudo-ground truth in the next training iteration.
However, the errors  of an immature classifier can make the process drift, usually introducing many of false positives in the training dataset. 
To alleviate this problem, we propose in this paper a training protocol based on the self-paced learning paradigm. 
The main idea is to iteratively  select a subset of images and boxes that are the most reliable, and use them  for training. While in the past few years similar strategies have been  adopted for SVMs and other classifiers, 
we are the first showing that a self-paced approach can be used with deep-network-based classifiers in an end-to-end training pipeline. 
The method we propose is built on the fully-supervised Fast-RCNN architecture and can be applied to  similar  architectures which represent the input image as a bag of boxes. 
We show state-of-the-art results on 
Pascal VOC 2007, Pascal  VOC 2010 
 and ILSVRC 2013. On ILSVRC 2013 our results based on a low-capacity AlexNet network outperform even 
those  weakly-supervised approaches which are based on much higher-capacity networks.  

\end{abstract}

\begin{IEEEkeywords}
Weakly supervised learning, object detection, self-paced learning, curriculum learning, deep learning, 
training protocol.
\end{IEEEkeywords}}

\maketitle

\IEEEdisplaynontitleabstractindextext

%
\IEEEpeerreviewmaketitle

\section{Introduction}
\label{Introduction}

A well known problem in object detection is the fact that collecting ground truth data (i.e., object-level annotations) for training is usually much more time consuming and expensive than collecting image-level labels for object classification. This problem is  exacerbated 
 in the context of the current 
 deep networks, 
  which need to be trained  or ``fine-tuned'' using large amounts of data.
 Weakly-supervised techniques for object detection (WSD) can alleviate the problem by leveraging existing datasets which provide image-level annotations only.
 
 \begin{figure*}
	\centering
	\includegraphics[width=0.9\linewidth]{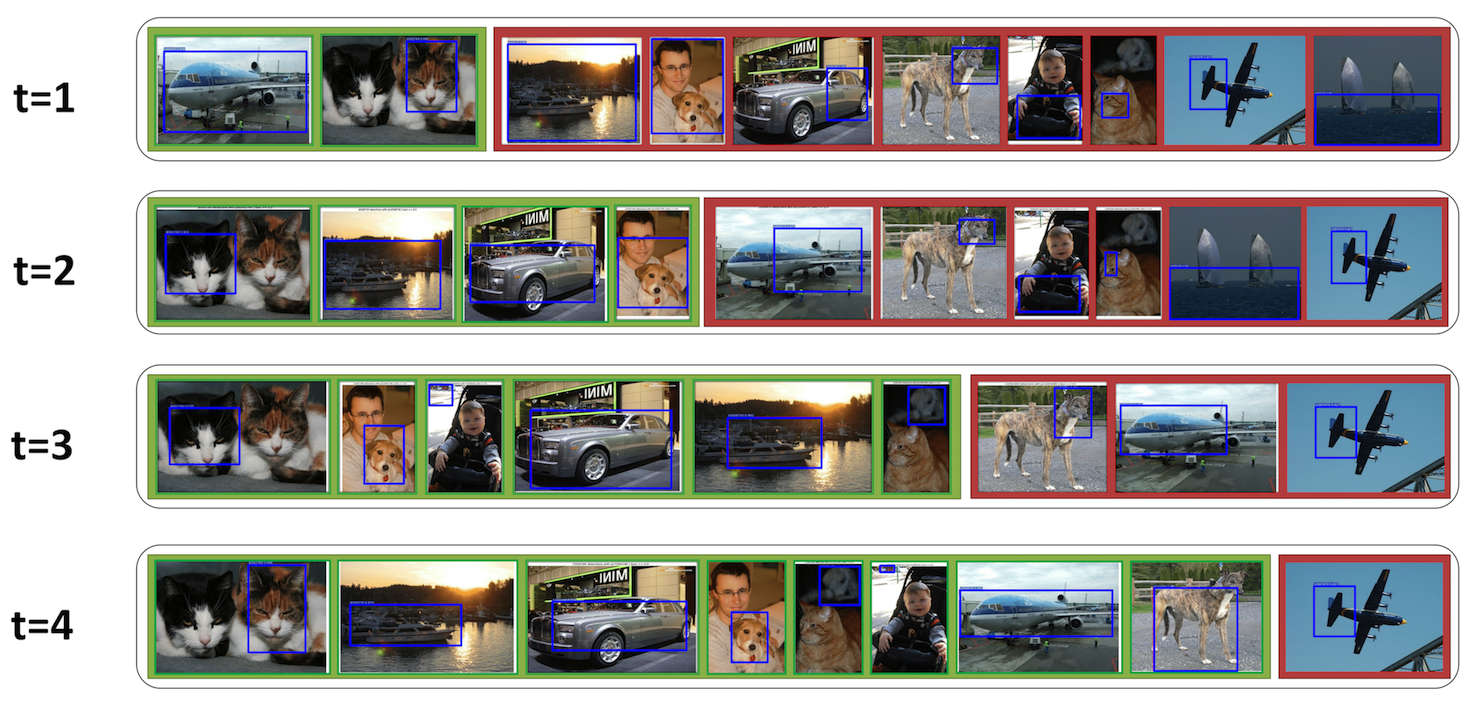}
	\caption{A schematic illustration of how the training dataset $T_t$ (represented by the green rectangle) of our deep network evolves depending on $t$ and on the progressively increasing recognition skills of the trained network.}
	\label{intro:teaser}
\end{figure*}

 In the common  Multiple Instance Learning (MIL) formalization of the WSD
 problem,
an image $I$, associated with a label of a given class $y$, is described  as a ``bag'' of
 Bounding Boxes (BBs), where at least one BB is a positive sample for  $y$ and the others are samples of the other classes (e.g., the background class). The main problem is how can the classifier, while being trained, automatically guess what  the positives in $I$ are.
A typical MIL-based solution 
alternates between 2 phases:  
(1) optimizing the classifier's parameters, assuming that the positive BBs in each image are known, and (2) using the current classifier to predict 
the most likely positives in each image \cite{DBLP:journals/pami/CinbisVS17,DBLP:journals/pr/NguyenTTR14}.
However, a well known problem of MIL-like solutions is that if the initial classifier is not strong enough, this process can easily drift. For instance, predicted 
false positives  (e.g., BBs on the background) can 
make the classifier learn something different than the target class.

Kumar et al. \cite{DBLP:conf/nips/KumarPK10} propose to alleviate this problem using a {\em self-paced} learning strategy. The main idea is that a subset of ``easy'' samples can be automatically selected by the classifier in each iteration. Training is then performed using only this subset, which is less prone to drifting and is progressively increased in the subsequent iterations when the classifier becomes more mature.
Self-paced learning, applied in many other studies \cite{DBLP:conf/nips/JiangMYLSH14,DBLP:journals/corr/LapedrizaPBT13,DBLP:conf/cvpr/LeeG11,DBLP:journals/corr/LiangLWLLY14,DBLP:conf/cvpr/PentinaSL15,DBLP:conf/cvpr/SupancicR13,DBLP:conf/ijcai/ZhangMZH16}, 
is related to {\em curriculum learning} \cite{DBLP:conf/icml/BengioLCW09} and is
biologically inspired by
the common human process of gradual learning, starting with the simplest concepts. 

In this paper we adopt a self-paced learning approach to handle the uncertainty related to the BB-level localization of the objects in the training images  in a WSD scenario, and ``easier'' is interpreted as ``more reliable'' localization.
We propose a new training protocol for deep networks in which the self-paced strategy is implemented by modifying the mini-batch-based  selection of the training samples. 
As far as we know, this is the first  self-paced learning approach  directly embedded in a modern end-to-end deep-network training protocol.

More specifically,
the solution we propose in this paper is based on the state-of-the-art (fully supervised) object-detection architecture Fast-RCNN \cite{DBLP:conf/iccv/Girshick15}. Fast-RCNN naturally embeds the idea of an image as a bag of BBs (see Sec.~\ref{FastRCNN}). Moreover, in the Fast-RCNN approach, 
  each mini-batch of the Stochastic Gradient Descent (SGD) procedure is sampled hierarchically, by first (randomly) sampling images and then sampling BBs within those images according to the BB-level ground truth information. We exploit this ``image-centric'' sampling but we modify the random image selection using a self-paced strategy in which the images containing the highest-confidence boxes associated with the annotated  classes are selected the first. In more detail, given an image $I$ with an image-level label $y$, we use the network trained in the previous iterations to  associate a class-specific score $s_{iy}$ with each {\em predicted} BB $p_i$. The highest-score box $z^I$ over all these predictions is selected in $I$. Note that, due to the spatial regression layer in Fast-RCNN, all the predicted BBs ($z^I$ included) are usually different from the set of input  box proposals, i.e., the bag of BBs associated with $I$ dynamically changes at every self-paced iteration.
  Once $z^I$ is chosen for each $I$ in the training set, we select a subset of images according to 
the score associated with the corresponding $z^I$  
  and a mini-batch of positive and background BBs is extracted using $z^I$.
Moreover, since we train a multi-class classifier (a common approach in deep networks which exploit inter-category representation sharing \cite{DBLP:conf/nips/KrizhevskySH12}), 
we exploit the competition among classifiers of different categories (i.e., among different classification-output neurons of the same network) and an image is chosen 
 only when its label is consistent with the strongest classifier on that image.
This image-based classifier competition is also used  to progressively train different classifiers starting from the  strongest ones. Since the predictions of the weak classifiers are usually 
not correct, we start training using only those samples corresponding to the strongest classifiers (i.e., the easiest classes), which are selected according to the number of images in which each classifier beats all the others.  The benefit of this strategy is that, during the initial training phases, the network learns a  visual representation of the objects in the shared layers (common to all the classes) together with a representation of the background class
and these improved representations are used in the subsequent training phases when  the network predicts the object localizations of the  difficult classes. 

Note that the basic Fast-RCNN architecture has been  used, simulated or extended in many different applications, due to the flexibility of its region-based pooling layer, including a few very recent  WSD architectures 
\cite{Bilen16,kantorov2016,Li_2016_CVPR}. 
Since we propose a {\em training protocol}, our method 
is orthogonal  
to many of these works 
and can potentially 
be used in conjunction with more sophisticated architectures to obtain higher accuracy experimental results.
However, we adopted the basic Fast-RCNN architecture \cite{DBLP:conf/iccv/Girshick15}
in order to present a more general  framework
 and we show empirical results using both a 
low-capacity 
AlexNet-like network \cite{DBLP:conf/nips/KrizhevskySH12} and a much larger VGG-16 network \cite{DBLP:journals/corr/SimonyanZ14a}.
In 
common WSD benchmarks (Pascal VOC 2007 and 2010, 
and ILSVRC 2013)
 our approach largely outperforms the current state of the art.
For instance, on ILSVRC 2013, our AlexNet-based results  are even higher than those results obtained by other  WSD methods which use much larger capacity networks 
(e.g., VGG-16). 

Finally, as far as we know, this is the first work empirically showing that a self-paced selection of samples is useful 
in training a deep network. Recent works have shown that an {\em anti-curriculum learning} strategy (e.g., hard-negative mining) can be useful in a supervised scenario 
\cite{Shrivastava_2016_CVPR,DBLP:conf/cvpr/SchroffKP15}. In \cite{DBLP:journals/corr/ZarembaS14,DBLP:conf/eccv/ShiF16} a {\em curriculum learning} strategy is used to select easy samples for training. However, the sample order is not decided by the classifier being 
trained (i.e., the network) but it is
provided using auxiliary  
data (e.g., the program complexity in \cite{DBLP:journals/corr/ZarembaS14} and the object's scale estimate in \cite{DBLP:conf/eccv/ShiF16}).
In our case, we do not  use auxiliary data and we assume that only image-level labels are given at training time. 
The goal of the adopted self-paced strategy is {\em to discard noisy training data} (wrong pseudo-ground truth BBs) and  we use the same detection network  that is being trained in order to progressively select the most likely candidate pseudo-ground truth BBs.

In summary, our contributions are the following.
\begin{itemize}
\item 
We propose a computationally efficient 
self-paced learning protocol  for training a deep network for WSD. During the training of the network, the same network, at different evolution stages, is used to predict the object-level localizations of the positive samples   and to select a subset of images whose pseudo-ground truth is the most reliable.
\item
We propose to use the spatial regression layer of the network to dynamically change the initial bag of boxes. We empirically show that selecting $z^I$ over the set of current predictions ($\{ p_i \}$) rather than over the set of the initial box proposals ($\{ b_i \}$) can largely boost the final WSD accuracy.
\item 
We propose to use class-specific confidence and inter-classifier competition to decrease the probability of selecting incorrect samples.
\item
We propose to extend  the self-paced sample selection paradigm to a self-paced class selection using the 
inter-classifier competition to estimate the difficulty of each class. 
\item 
We test our approach on Pascal VOC 2007, Pascal  VOC 2010 
 and ILSVRC 2013, obtaining state-of-the-art WSD results in all these benchmarks. This is the first work showing the usefulness of a self-paced sample selection strategy with an end-to-end trained deep network.
\end{itemize}
 
Our code   and our trained models
are publicly available\footnote{\url{https://github.com/moinnabi/SelfPacedDeepLearning}}.  
 The rest of the paper is organized as follows. In Sec.~\ref{RelatedWork} we review the literature concerning self-paced learning, weakly supervised object detection and related areas. In  
 Sec.~\ref{FastRCNN} we analyse the main aspects of the Fast-RCNN architecture that are of interest for our work. In Secs.~\ref{WeaklySelf-Paced}-\ref{Discussion}
  we introduce and analyse our method, which is evaluated in Secs.~\ref{WeaklySupervisedObjctDetection}-\ref{Analysis} 
  and finally we conclude in  
 Sec.~\ref{Conclusions}.

\section{Related work}
\label{RelatedWork}

Many recent studies have shown that selecting a subset of ``good'' samples for training a classifier can lead to better results than using all the samples 
\cite{DBLP:journals/corr/LapedrizaPBT13,DBLP:conf/cvpr/LeeG11,DBLP:conf/eccv/Sangineto14,DBLP:conf/cvpr/SupancicR13,DBLP:journals/tip/WangHRZM15}. 
A pioneering work in this direction is the {\em curriculum learning} approach proposed in \cite{DBLP:conf/icml/BengioLCW09}. The authors show that suitably sorting the training samples, from the easiest to the most difficult, 
and iteratively training a classifier starting with a subset of easy samples (progressively augmented with more and more difficult samples), can 
be useful to find better local minima.  
In \cite{DBLP:journals/corr/ChenG15a}, easy and difficult images (taken from datasets known to be more or less ``difficult'') are provided for training a Convolutional Neural Network (CNN) in order to learn  generic CNN features using webly annotated data. 
In \cite{DBLP:journals/corr/WeiLCSCZY15}, different and progressively more complex CNNs are trained for a segmentation task, using more and more difficult data samples together with the output of the previously learned networks.
It is worth noting that in these and in all the other curriculum-learning-based approaches, 
the order of the samples is decided using  additional supervisory information 
usually provided by a human teacher.
Unfortunately, these ``image-easiness'' meta-data are not available for the common large-scale datasets.

Curriculum learning was extended to {\em self-paced} learning in \cite{DBLP:conf/nips/KumarPK10}. The main difference between the two paradigms is that in self-paced learning the order of the samples is automatically computed and it is a priori unknown. 
 The selection of the best ``easy'' sample set for training is, generally speaking, untractable  (it is a subset selection problem). The solution proposed in \cite{DBLP:conf/nips/KumarPK10}  is based on a continuous relaxation of the problem's constraints which leads to a biconvex  optimization of a Structural SVM.
Supancic  et al. \cite{DBLP:conf/cvpr/SupancicR13} adopt a similar framework 
in a tracking by detection scenario
and train a detector using a subset of video frames, showing that this selection is important to avoid drifting. Frames are selected by computing the SVM objective function for different candidate subsets of frames and then selecting the subset corresponding to the lowest objective value.
In \cite{DBLP:conf/nips/JiangMYLSH14} the authors pre-cluster the training data in order to balance the selection of the easiest samples with a sufficient inter-cluster diversity. 
However, the clusters and the feature space are fixed: they do not depend on the current self-paced training iteration 
and the adaptation of this method to a deep-learning scenario, where the feature space changes during learning, is not trivial.
In \cite{DBLP:conf/cvpr/PentinaSL15} a set of learning tasks is automatically sorted  in order to allow for a gradual sharing of information among tasks. Our self-paced {\em class} selection aims at a similar goal but it is obtained with a radically different approach.
Liang \textit{et al.} \cite{DBLP:journals/corr/LiangLWLLY14}  use Exemplar SVMs (ESVMs) \cite{malisiewicz-iccv11} to train a classifier from a single positive sample. The trained ESVMs are then run on an unsupervised collection of videos in order to extract new positives which are gradually  more and more  different from the seed instances. ESVMs are also used in \cite{DBLP:journals/corr/LapedrizaPBT13} to assess the ``training value'' of each instance and then use this value to select the best subset of samples for training a classifier. In \cite{DBLP:conf/cvpr/LeeG11}, the easiness of an  image region  is estimated using its ``objectness'' and the category context of its surrounding regions. In \cite{DBLP:conf/ijcai/ZhangMZH16} saliency is used to progressively select samples in WSD.

Although some of these self-paced methods use pre-trained CNN-based features to represent samples (e.g., \cite{DBLP:conf/nips/JiangMYLSH14,DBLP:journals/corr/LiangLWLLY14}), none of them 
uses a deep network as the classifier or formulates the self-paced strategy in an end-to-end deep-network training protocol as we do in this paper.

Concerning the broader WSD field, 
a few recent studies address the problem in a deep-learning framework. For instance, in \cite{DBLP:conf/cvpr/OquabBLS15}, a
 final max-pooling layer selects the highest scoring position for an instance of an object in the input image and back-propagates the training error only to those  network's weights that correspond to the highest scoring window. However, in this work the object is localized at testing time by providing  only one 2D point. 
  A similar max-pooling 
  layer over different subwindows of the input image is adopted in \cite{DBLP:journals/corr/GkioxariGM15}, together with the Fast-RCNN architecture \cite{DBLP:conf/iccv/Girshick15}, to select the most significant context box in an action recognition task.
Hoffman \textit{et al.} \cite{DBLP:journals/corr/HoffmanGTDGDS14} use both weakly-supervised and strongly-supervised data (the latter being BB-level annotations) to adapt a CNN pre-trained for a classification task to work in a detection task. This work was extended in \cite{DBLP:conf/cvpr/HoffmanPDS15} using a MIL-based SVM training. Encouraging results were obtained both in \cite{DBLP:journals/corr/HoffmanGTDGDS14} and in  \cite{DBLP:conf/cvpr/HoffmanPDS15} using the ILSVRC 2013 detection dataset. 
  However, in both papers, auxiliary strongly-annotated data for half of the 200 ILSVRC 2013
  categories were used for training, together with image-level-only annotations for the remaining categories.
  
Very recently, a few WSD approaches have been proposed for training a deep network in an end-to-end fashion which are based  on 
specific network architectures. 
For instance, Bilen and Vedaldi \cite{Bilen16} extend a Fast-RCNN-like network 
using  two  different data streams, respectively computing a classification and a detection score for each candidate box of an image. 
Specifically, the detection score  is obtained using a softmax operator which produces a probability distribution over all the input region proposals, thus avoiding the  hard assignment of the  pseudo-ground truth position to a specific box, common in MIL-like approaches. A similar soft assignment in a WSD scenario was previously developed in \cite{Bilen1,DBLP:conf/cvpr/BilenPT15}, while
the architecture proposed in \cite{Bilen16} is further extended in \cite{kantorov2016} introducing specific regions which describe the context surrounding each candidate box.
We compare with \cite{Bilen1,DBLP:conf/cvpr/BilenPT15,Bilen16} and \cite{kantorov2016} in 
Sec.~\ref{WeaklySupervisedObjctDetection}.

Li et al.  \cite{Li_2016_CVPR} address the multi-label problem (the same image can contain  objects belonging to different classes) by proposing a specific classification loss for training an image {\em classification} network.  Then, this classification network is used to initialize a Fast-RCNN-based {\em detector} which is trained using a MIL framework.  Following their  approach we trained a 
similar 
 classification network which is used as the initialization of our detector, trained using our self-paced framework and tested
 on Pascal VOC, where we largely outperform the results obtained in \cite{Li_2016_CVPR}. In \cite{attentionBMVC2016} for each candidate box of an image, an {\em attention score} is computed which estimates how likely that box contains the object of interest. While these works propose specific network {\em architectures} for solving the WSD problem, we take a different direction and we propose a {\em training protocol} which can be used with different architectures, provided that they have a region-pooling layer similar to Fast-RCNN 
and an image-based sampling strategy in computing the SGD mini-batch 
 (Sec.~\ref{FastRCNN}). 
 
 Finally, the closest work to this paper is probably 
 \cite{DBLP:conf/eccv/ShiF16}, where the authors propose a curriculum-learning based training protocol for WSD, in which the  size estimate of an object inside a given image is used as a proxy for assessing the ``easiness'' degree of that image.  
  However, an additional training set, provided with the ground-truth size of each object, is necessary to train the size regressor, which makes this approach not directly comparable with other works using only weakly-supervised data. Moreover, the authors present results using either SVM classifiers or a deep network (Fast-RCNN). In the latter case, the deep network is trained using a simpler 
  MIL approach in which 
 the previous SVM-self paced based image selection is used only to select an initial set of pseudo-ground truth 
for training. Differently from  \cite{DBLP:conf/eccv/ShiF16}, our pseudo-ground truth training set is modified during the Fast-RCNN training.

\section{Fast-RCNN and notation}
\label{FastRCNN}

In this section we review the main aspects of the Fast-RCNN \cite{DBLP:conf/iccv/Girshick15} approach which are important to understand our proposal and we introduce  notations, used in the rest of the paper.

The {\em supervised} state-of-the-art object-detection approaches \cite{DBLP:conf/iccv/Girshick15,Shrivastava_2016_CVPR,DBLP:conf/iccv/HeGDG17} on Pascal VOC \cite{pascal-voc-2007,pascal-voc-2010} are based on the
  Fast-RCNN architecture,
whose main characteristic is the {\em RoI pooling layer}. This layer is used to extract box-specific information from the final convolutional maps and feed the final classification and regression branches of the network.

The network takes as input an image $I$ (raw pixels) and a set of BBs on $I$: 
$B(I) = \{ b_1, ..., b_n \}$ . $B(I)$ is  computed using an external tool, which usually selects image subwindows taking into account their ``objectness'': for instance using Selective Search \cite{DBLP:journals/ijcv/UijlingsSGS13} (also used in all our experiments). 
If $f$ is the function computed by the network, 
its outcome is a set of detections:

\begin{equation}
\label{eq.fast-rcnn}
f(I, B(I)) = \{ d_{ic} \}_{i= 1,...,n, c= 1,...,C},
\end{equation}

\noindent
where $C$ is the number of object classes and, for each class $c$ and each input box $b_i \in B(I)$,
  $d_{ic} = (s_{ic}, p_{ic})$, where $s_{ic}$ is the score and $p_{ic}$ the predicted box. Note that, usually, $p_{ic} \neq b_i$ and $p_{ic} \not \in B(I)$,  $p_{ic}$ being the result of a spatial regression  applied to $b_i$.
The RoI pooling layer makes it possible to efficiently compute $f(I, B(I))$ and the dependence of the network's output on a set of boxes $B(I)$ is important for our bag of BBs formulation. 

As mentioned in Sec.~\ref{Introduction}, another aspect of Fast-RCNN  exploited in our training protocol is that each mini-batch (used in the mini-batch SGD procedure) is constructed using only a small number $m$ of images, where $m = 2$ is indicated as a good compromise between  quality of the samples and efficiency. Specifically, at training time  a  set  
$T = \{ (I_1, G_1), ..., (I_j, G_j) ..., (I_N, G_N) \}$ of $N$ images and corresponding ground-truth is  given, where: $G_j = \{ (y_1, b_1), (y_2, b_2), ... \}$ and, for each $(y_i, b_i)$, 
   $y_i \in \{1, ..., C \}$ is the label and $b_i \in B(I)$ is the BB of the i-th object instance in $I$.
In each SGD iteration $m = 2$ images are randomly extracted from $T$. If $I_j$ is one of these 2 images, for each $(y_i, b_i) \in G_j$, $b_i$ is matched with the boxes in $B(I)$ using common spatial criteria (i.e., Intersection over Union  
between two BBs higher than a given threshold) in order to select those BBs in $B(I)$ that will be used as positives for the $y_i$ class, as well as a set of ``negatives'' (i.e., samples for the background class $y = 0$).

For more details  we refer the reader to \cite{DBLP:conf/iccv/Girshick15}.
What is important to highlight here is that  Fast-RCNN is a strongly supervised method. Conversely, in our weakly-supervised scenario, we do not have BB-level annotations.
Hence, in the rest of the article we assume that our training set is 
$T = \{ (I_1, Y_1), ..., (I_j, Y_j), ..., (I_N, Y_N) \}$, 
where $Y_j = \{ y_1, ... y_{n_j} \}$ is the set of labels associated with image $I_j$ and the number of object classes contained in $I_j$  varies depending on the specific image ($n_j \geq 1$). Note that, for a given class $y_i \in Y_j$, more than one object of the same class can be contained in $I_j$: for instance, two instances of the ``dog'' 
category; and the number of instances is unknown.

Since object-level ground truth is not given, we use the network (in the current self-paced training iteration) to compute the most likely positions of the objects in $I_j$. 
In the next section we show how these locations are computed and how $T$ is updated following a self-paced learning strategy.

\section{Self-paced learning protocol}
\label{WeaklySelf-Paced}

We call $W$ the set of weights of all the layers of the network and we initialize our network with $W_0$, which can be obtained using any standard object {\em classification} network, {\em trained using only image-level information}. 
At the end of this section 
we provide more details on how $W_0$ is obtained. 

The proposed self-paced learning protocol 
of the network is composed of a sequence of {\em self-paced iterations}. At a self-paced iteration 
$t$ we use the current network $f_{W_{t-1}}$ in order to select a subset of easy classes and easy samples of these classes. The result is a new training set $T_t$ which is used 
 to train a new model $W_t$. $W_t$ is obtained using the ``standard'' training procedure of the Fast-RCNN (Sec.~\ref{FastRCNN}), based on mini-batch SGD, but it is applied to $T_t$ {\em only} 
and iterated for only  $N_t/m$
 mini-batch SGD iterations, $N_t$ being the cardinality of $T_t$. 
Note that, being $m$ the number  of images used to build a mini-batch,  $N_t/m$ corresponds to one epoch (a full iteration over $T_t$).
 Note also that a mini-batch SGD iteration is different from a self-paced iteration and in each SGD iteration a mini-batch of BBs is formed using the pseudo-ground truth obtained using $f_{W_{t-1}}$.
The proposed protocol is summarized in Alg.~\ref{alg.SelfPaced.training} and we provide the details  below.

\begin{algorithm}[t]
  \caption{Self-Paced Weakly Supervised Training}
 \begin{algorithmic}
  \State
  \textbf{Input:} $T$, $W_0$, $r_1$, $M$ 
  \State
  \textbf{Output:} Trained network $f_{W_M}$
   \State
\hspace{3pt} 1 For $t := 1$ to $M$: \State
\hspace{3pt} 2 \indent $P:= \emptyset$, $T_t:= \emptyset$ \State
\hspace{3pt} 3 \indent For each $(I,Y) \in T$: \State
\hspace{3pt} 4 \indent \indent Compute $(s_y^I, z_y^I)$ using Eq.~\ref{latent-box} \State
\hspace{3pt} 5 \indent \indent If $y \in Y$, then: $P:= P \cup \{ (I, s_y^I, z_y^I, y) \}$ \State
\hspace{3pt} 6 \indent  For each $c \in \{ 1, ..., C \}$ compute $e(c)$ using Eq.~\ref{class-easiness} \State 
\hspace{3pt} 7 \indent  $C_t := r_t C$ \State 
\hspace{3pt} 8 \indent  Let $S = \{ c_1, c_2, ... \}$ be the subset of the $C_t$ easiest \State 
\indent \indent    classes according to  $e(c)$ \State 
\hspace{3pt} 9 \indent Remove from $P$ those tuples $(I, s, z, y)$ s.t. $y \not\in S$ \State
10 \indent  $N_t: = min(r_t N, |P|)$ \State 
11 \indent  Let $P'$ be the $N_t$ topmost tuples in $P$ according to \State
\indent \indent the $s$-score \State
12 \indent  For each $(I, s, z, y) \in P'$: $T_t:= T_t \cup \{ (I, \{ (y, z) \} ) \}$ \State
13 \indent  $V_0 = W_{t-1}$ \State
14 \indent  For $t' := 1$ to $N_t/m$: \State
15 \indent \indent Randomly select  \State
\indent \indent \indent $(I_1, \{ (y_1, z_1) \} ), ..., (I_m,  \{ (y_m, z_m) \}) \in T_t$ \State
16 \indent \indent Compute a mini-batch $MB$ of BBs using \State
\indent \indent \indent $(I_1, \{ (y_1, z_1) \} ), ..., (I_m,  \{ (y_m, z_m) \})$ \State
17 \indent \indent Compute $V_{t'}$ using $MB$ and \State
\indent \indent \indent back-propagation on $f_{V_{t'-1}}$ \State 
18 \indent $W_t:= V_{N_t/m}$ \State
19 \indent  $r_{t+1} = r_t + \frac{1-r_1}{M}$ \State
 \end{algorithmic}
 \label{alg.SelfPaced.training}
\end{algorithm}

\bigskip
\noindent
{\bf Computing the latent boxes.}
 Given an image $I$, its label set $Y$ and the current network $f_{W_{t-1}}$, first we compute:

\begin{equation}
  (s_y^I, z_y^I)  = \argmax_{  (s_{ic}, p_{ic}) \in f_{W_{t-1}}(I, B(I))} s_{ic} 
  \label{latent-box} 
\end{equation}

\noindent
In Eq.~\ref{latent-box}, $(s_y^I, z_y^I)$ is the detection in $f(I, B(I))$ with the highest score  ($s_y^I$) with respect to all the detections obtained starting from  $B(I)$ and the subscript $y$ indicates the corresponding class. 
$z_y^I$ is a {\em latent box} which specifies the most likely  position of an object of the ``winning'' class $y$ in image $I$ according to $f_{W_{t-1}}$. Note that the background class is not included in 
$f_{W_{t-1}}$ (see Eq.~\ref{eq.fast-rcnn}), thus $y \in \{1, ..., C \}$.
Note also that $z_y^I$ is computed using the regression layer of $f_{W_{t-1}}$ and usually 
$z_y^I \not \in B(I)$. Conversely, most of the existing WSD approaches \cite{Bilen16,kantorov2016,Li_2016_CVPR,attentionBMVC2016,DBLP:conf/eccv/ShiF16} restrict the choice of the pseudo-ground truth over a pre-fixed and constant set of region proposals ($B(I)$).  We empirically show in Sec.~\ref{Ablation} the importance of using the regression part of the network  to extend $B(I)$. However, at the end of the current iteration $t$, $z_y^I$ and all the other predictions $\{p_{ic}\}$ are discarded and, at iteration $t+1$, ~Eq.~\ref{latent-box} is minimized again starting from the original set of box proposals $B(I)$. 

If $y \in Y$ ($y$ is one of the labels of $I$), then we associate $I$ with the latent box $z_y^I$ and with the confidence score $s_y^I$
 (line 5), 
 otherwise $I$ is temporally discarded and will not be included in the current self-paced iteration  training set $T_t$.

 \begin{figure}
	\centering
	\includegraphics[width =\linewidth]{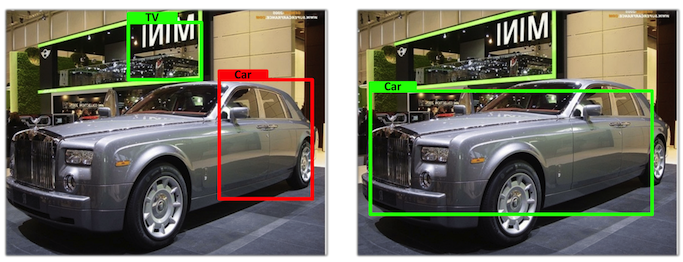}
	\caption{Inter-classifier competition. The two above figures show the behaviour of our self-paced learning algorithm  on the same image $I$ in two different iterations ($t = 1$ and $t = 2$). In both cases, the green box shows the highest-score box in $I$ corresponding to $z_y^I$ in Eq.~\ref{latent-box}. Conversely, the red box in the left figure shows the highest-score box in $I$ corresponding to the $car$ class ($z_{car}$). Since in $t = 1$ (left figure) $s_{TV} = s_y^I > s_{car}$, and since $TV$ $\not\in Y$, then $I$ is not included in $T_1$ (see also Fig.~\ref{intro:teaser}). However, in 
	$t = 2$  (right figure) $s_{TV} < s_y^I = s_{car}$ (where $car \in Y$), thus $(I, s_{car}, z_{car}, car)$ is included in $P$ (line Line 5 
 in Alg.~\ref{alg.SelfPaced.training}): the ``car'' classifier in this iteration is strong enough 
and 	 ``wins'' in $I$.}
	\label{comp.fig}
\end{figure}

\bigskip
\noindent
{\bf Inter-classifier competition.} 
Eq.~\ref{latent-box} imposes a competition among classifiers, where a 
``classifier'' 
 for  class $c$ is the 
classification-output neuron of $f$ specific for class $c$. 
Only if one of the classifiers {\em corresponding to an image label} $y \in Y$ is ``stronger'' (more confident) than all the others, including $y' \not \in Y, y' \neq 0$,  then $I$ 
is considered for  
 inclusion in  
 $T_t$
(Line 5). 
We found this competition to be very important 
to decrease the risk of error and to enforce a self-paced learning  strategy which prudently selects initially easy image samples. When the network becomes more mature (i.e., in the subsequent self-paced iterations), the risk of error gradually decreases  and a previously weaker classifier can correctly ``win'' a previously discarded image (see Fig.~\ref{comp.fig}). 

Note that a consequence of this classifier competition is that only one pseudo-ground truth box $z_y^I$ can be selected from a given image $I$, regardless of the number of labels associated with $I$ and the number of object instances of class $y$ in $I$. In Sec.~\ref{SIML-MIML} we present multiple-label and multiple-instance relaxations of this inter-classifier competition.

\bigskip
\noindent
{\bf Class selection.}
The classifier competition is used also to sort all the $C$ classes from the easiest to the most difficult. 
This is obtained using the winning classifiers in each image as follows. Let 
$P = \{ (I_1, s_1, z_1, y_1), (I_2, s_2, z_2, y_2), ... \}$ be the set of non-discarded images at iteration $t$ (Line 5) 
and let:

\begin{equation}
\label{class-easiness}
e(c) = | \{ (I, s, z, y) : (I, s, z, y) \in P, y = c \}| / p_c
\end{equation}

\noindent
be the ``easiness'' degree of class $c$, defined as the ratio between
the cardinality of those tuples in $P$ associated with the class label $y = c$ over the overall frequency of the label $c$ in $T$ ($p_c$). 
The higher the value $e(c)$ for a given class $c$, the stronger
the corresponding classifier is and the easier that class according to
 $f_{W_{t-1}}$. We sort all the classes using  $e(c)$ ($c = 1, ..., C$) and we select the subset of the easiest $r_t C$ classes which are the only classes subject to training in the current self-paced iteration. The ratio $r_t \in [0,1]$ is  increased at each self-paced iteration (see below) and at iteration $t+1$ more difficult classes will be included in $T_{t+1}$ and presented  
to the network for training.

\bigskip
\noindent
{\bf Selecting the easiest image samples.} Once image samples associated  with difficult classes have been removed from $P$ (Line 9), 
  we select a subset ($T_t$) of $P$ corresponding to those images in which  $f_{W_{t-1}}$ is the most confident. With this aim we use the score $s_y^I$ computed using Eq.~\ref{latent-box} and we sort $P$ in a descending order according to these scores. Then we select the first $N_t$ top-most elements, where $N_t =  min(r_t N, |P|)$, and $r_t N$  is an upper bound on the number of elements of $T$ to be selected in the current self-paced iteration. At each self-paced iteration $r_t$ is  increased. Indeed, we adopt the strategy proposed in
\cite{DBLP:conf/nips/KumarPK10} (also used in most of the self-paced learning approaches) to progressively increase the training set as the model is more and more mature (see Fig.~\ref{intro:teaser}).
However, in our experiments we observed that  usually $|P| < r_t N$, mainly because of the sample rejection step in Line 5, 
hence the learning ``pace'' is dominated by our classifier-competition constraint.

\bigskip
\noindent
{\bf Details.}
The inner loop 
over $t'$ 
(Lines 14-17), 
 whose number of iterations depends on the length of the current training set $T_t$, is equivalent to the mini-batch SGD procedure adopted in \cite{DBLP:conf/iccv/Girshick15}, with a single important difference: Since we do not have BB-level ground truth, each mini-batch is computed using $(y, z)$ as the pseudo-ground truth (Line 12). 
$MB$ is built using BB samples which are collected   using the same spatial criteria adopted in the supervised Fast-RCNN training protocol (see Sec.~\ref{FastRCNN}) and with the same positive/negative proportion (see \cite{DBLP:conf/iccv/Girshick15} for more details). 
Also the number of images $m = 2$ we use to compute a mini-batch of BBs is the same used in \cite{DBLP:conf/iccv/Girshick15}.
In this  loop, the weights of the network are called $V_{t'}$ for notational convenience (their update depends on $t'$ and not on $t$), but there is only one network model, continuously evolving. 

Inspired by \cite{DBLP:conf/nips/KumarPK10}, where half of the data are used in the first self-paced iteration and all data are used in the last iteration, we start with $r_1 = 0.5$ and 
we iterate for $M = 4$ iterations till $r_M = 1$, linearly interpolating the intermediate increments (Line 19). 
Experiments with $M = 5$ obtained very similar  results.
All the other Fast-RCNN specific hyper-parameters  are the same used in \cite{DBLP:conf/iccv/Girshick15} for the fine-tuning of a pre-trained network, including the initial learning rate value (0.001), the size of $MB$ (128), the weight decay (0.0005) and the momentum (0.9). No batch normalization is used and standard backpropagation with momentum is adopted. The only difference with respect to \cite{DBLP:conf/iccv/Girshick15} is that, in all our experiments,  we divide (only once) the learning rate by a factor of 10 after the first  self-paced iteration.

The reason for which we adopted the same  hyper-parameter values used in the supervised Fast-RCNN 
and we followed as strictly as possible the same design choices (e.g., how a mini-batch is computed, etc.)
is that tuning the hyper-parameter values in a weakly supervised scenario is not easy because of the lack of validation data with BB-level ground truth. Moreover, in this way our training protocol can be more easily generalized to those WSD approaches which are based on the same Fast-RCNN architecture. 
All the above hyper-parameters (including those which are specific of our self-paced protocol, $r_1$ and $M$) are kept fixed in 
 our experiments on both Pascal VOC and  ILSVRC. 
 
 Finally, in all our experiments, $T$ is composed of the original images and their mirrored versions. No other data augmentation is performed.

\bigskip
\noindent
{\bf Computational issues.} From a computational point of view, the only 
additional demanding operation in our approach with respect to the Fast-RCNN training procedure is 
computing $f(I, B(I))$ for each $I \in T$, which involves passing $I$ forward through all the  layers of $f$. 
Fortunately, Fast-RCNN performs this operation in only $\approx 0.1$ seconds per image (e.g., using a Tesla K40 GPU).
For instance, with $N = 20K$, computing the latent  boxes of all the images in $T$ takes approximately 30  minutes.
Note that this operation is repeated only $M$ times during the whole training.

A simpler self-paced   approach to train a Fast-RCNN is to {\em fully} train the network 
(for several epochs) with an initial, small ``easy'' dataset $T_1$, then use the current network to compute the latent variables of a larger set $T_2$, then {\em fully} fine-tune the network again, etc. However, this procedure is not only much slower than the epoch-based dataset update strategy we adopted (because it involves a full-training of the detector for each iteration), but it is also less effective. Our preliminary results using this approach showed that the network quickly overfits
on the initial relatively small dataset $T_1$ and the final accuracy of the network is  much lower  than what we obtain using Alg.~\ref{alg.SelfPaced.training}.

\bigskip
\noindent
{\bf Initialization.} The initial model $W_0$ can be obtained in different ways using only weakly-supervised annotation.
Below we describe  the steps we followed in our experiments on Pascal VOC and on ILSVRC as   solution examples.

In both datasets we used a two-steps procedure: (1) training a Classification Network (CN) and (2) 
inspired by \cite{DBLP:conf/nips/KumarPK10}, where {\em all} the samples of the 
 dataset are used for pre-training the classifier using a small number of iterations, we also 
pre-train the Detection Network (Fast-RCNN) using all the images of $T$ for a fixed, small number of SGD iterations.

In case of the ILSVRC dataset, the CN is obtained following the steps suggested in
 \cite{DBLP:journals/corr/HoffmanGTDGDS14,DBLP:conf/cvpr/HoffmanPDS15}: 
Starting from the  AlexNet \cite{DBLP:conf/nips/KrizhevskySH12}, pre-trained on ImageNet (1000 classes), we first fine-tune the network on the
ILSVRC 2013 
detection dataset \cite{DBLP:journals/corr/RussakovskyDSKSMHKKBBF14}, which is composed of $C = 200$ classes. This is done by removing the last layer from the AlexNet and replacing it with a 200-class output layer. For the fine-tuning we use a random subset of the {\em train} partition of ILSVRC 2013
 (see Sec.~\ref{WeaklySupervisedObjctDetection}), but we simulate a situation in which we have access {\em to image-level labels only}. We call $h^I$ this CN and $W_{CN}^I$  its weights, where the superscript $I$ stands for ``trained on ILSVRC 2013''. Note that $h^I$ takes as input a $227 \times 227$  image and outputs a 200-element score vector.
 
Using $W_{CN}^I$ we initialize the Fast-RCNN architecture. To do so, the last layer needs  again be removed and replaced by two (parallel) Fast-RCNN specific layers: a $C + 1$ classification layer and $C \times 4$ 
regression layer \cite{DBLP:conf/iccv/Girshick15}. The weights of this layers are randomly initialized. Then, we train the Fast-RCNN Detection Network (DN) for $30K$ SGD iterations 
using all the images in $T$, where $T$ is the {\em val1} split of ILSVRC 2013 (see Sec.~\ref{WeaklySupervisedObjctDetection}), mirrored images included. Since Fast-RCNN is a DN and needs BB-level annotation for training, we associate the images in $T$ with 
 a pseudo-ground truth  by collecting  the top-score boxes obtained using $h^I$.
More in detail, for each  $(I,Y) \in T$ and each box $b \in B(I)$, we rescale $b$ 
to $227 \times 227$ 
($u(b)$) and, for each $y \in Y = \{ y_1, ..., y_k \}$, we compute:

\begin{equation}
\label{MIL1}
 z_y  = \argmax_{  b \in B(I)} h^I(u(b),y),
\end{equation}

\noindent
where $h^I(\cdot,y)$ is the $y$-class score. The final 
pseudo ground-truth   corresponding to $I$ is $G = \{ (y_1, z_{y_1}) ..., (y_k, z_{y_k}) \}$
 (see Sec.~\ref{FastRCNN}). We call this training protocol $Init$ and you can think of it as a one-shot MIL solution with only one iteration over the latent variables (i.e., the latent boxes are not recomputed while the network is trained and are kept fixed).
Note that there is no classifier competition or class selection
 in $Init$  and we use all the samples in $T$, 
 inspired by Kumar et al.  \cite{DBLP:conf/nips/KumarPK10}, 
 confirming that this strategy leads to a good initialization for a self-paced learning approach. At the end of $Init$, we call the final network's weights $W_0^I$ and we use $W_0^I$ as input 
 in Alg.~\ref{alg.SelfPaced.training}.

In case of Pascal VOC we use a similar strategy and we train  different CNs for different experiments, using as basic architecture either AlexNet or VGG-16 \cite{DBLP:journals/corr/SimonyanZ14a}.
Firstly, we simply use the above-mentioned $h^I$ {\em trained on ILSVRC 2013}: the weights ($W_{CN}^I$) of $h^I$ are directly used to initialize the Fast-RCNN architecture (except the last randomly initialized layers, see above). Moreover, the pseudo-ground truth for the $Init$  stage is collected using Eq.~\ref{MIL1} with an amputated version of $h^I$,   obtained   
by removing 180 over 200 output neurons and keeping only those $fc8$ original neurons roughly corresponding to the 20 Pascal VOC classes\footnote{Specifically, the adopted 
ILSVRC $\rightarrow$ VOC class mapping is: 
{\em airplane}  $\rightarrow$ {\em aeroplane}, 
{\em bicycle}  $\rightarrow$ {\em bicycle},
{\em bird}  $\rightarrow$ {\em bird},
{\em watercraft}  $\rightarrow$ {\em boat},
{\em wine bottle}  $\rightarrow$ {\em bottle},
{\em bus}  $\rightarrow$ {\em bus},
{\em car}  $\rightarrow$ {\em car},
{\em domestic cat}  $\rightarrow$ {\em cat},
{\em chair}  $\rightarrow$ {\em chair},
{\em cattle}  $\rightarrow$ {\em cow},
{\em table}  $\rightarrow$ {\em diningtable},
{\em dog}  $\rightarrow$ {\em dog},
{\em horse}  $\rightarrow$ {\em horse},
{\em motorcycle}  $\rightarrow$ {\em motorbike},
{\em person}  $\rightarrow$ {\em person},
{\em flower pot}  $\rightarrow$ {\em pottedplant},
{\em sheep}  $\rightarrow$ {\em sheep},
{\em sofa}  $\rightarrow$ {\em sofa},
{\em train}  $\rightarrow$ {\em train},
{\em tv or monitor}  $\rightarrow$ {\em tvmonitor}.}. 
We run $Init$   for $10K$ SGD iterations
using  the {\em trainval} split of Pascal VOC 2007 (see Sec.~\ref{WeaklySupervisedObjctDetection}) as our $T$. 
 We call $W_0^{I,P}$ the final network's weights 
   because they are obtained using a hybrid solution, based on training images from both ILSVRC 2013 and Pascal  VOC 2007.
Note that, due to the differences between ILSVRC  and Pascal VOC in both the corresponding {\em image distributions} and the {\em object classes} 
\cite{Bazzani:WACV16,DBLP:conf/cvpr/GirshickDDM14,ECCV12_Khosla}, 
 using $h^I$ to initialize $Init$ corresponds to a quite weak initialization. Despite this, our experimental results show a surprisingly good accuracy 
achieved after the proposed self-paced training procedure (see Sec.~\ref{WeaklySupervisedObjctDetection}).

In case of Pascal VOC, we also fine-tune a second CN, using only Pascal VOC 2007 {\em trainval}.  Also in this case the basic network architecture is  AlexNet, pre-trained on ImageNet (1000 classes). However, since Pascal VOC 2007 {\em trainval} is a much smaller dataset than the ILSVRC 2013 {\em train} split and, on average, a Pascal VOC image contains 
more objects with different-labels  than an ILSVRC 2013 image 
\cite{DBLP:conf/cvpr/GirshickDDM14}, 
care should be taken in training a CN directly on Pascal VOC. For this reason we adopted the approach proposed in \cite{Li_2016_CVPR} for training a CN on a multi-label dataset, where the authors replace the network {\em softmax} loss with a multi-label loss based on a 
$2C$ binary element  
vector label.
The trained CN ($h^P$) and the corresponding weights ($W_{CN}^P$) are used to collect pseudo-ground truth data and to initialize the Fast-RCNN for the $Init$  stage (see above), using the same number ($10K$) of  SGD iterations and the final weights are called $W_0^P$.
Finally, in Sec.~\ref{WeaklySupervisedObjctDetection} 
we also show two  experiments (Tab.~\ref{tab.Pascal07-VGG} and Tab.~\ref{tab.Pascal10-VGG16}) in which the basic architecture is VGG-16 and the 
initialization procedure is the same followed in case of $W_0^P$. To simplify our notation, we call the VGG-16 based initialization $W_0^P$ 
as well 
and we will explicitly specify when the basic architecture is not AlexNet.

\section{Discussion}
\label{Discussion}

As mentioned in Sec.~\ref{Introduction}, the goal of the proposed  self-paced strategy is to discard noisy data in order to decrease the drifting problem in a MIL framework. Hence, the training protocol described in  Alg.~\ref{alg.SelfPaced.training} can be seen as a ``prudent'' strategy, where ``good'' (``easy'') data are preferred to lot of data. Moreover, Alg.~\ref{alg.SelfPaced.training} can terminate without using all $T$. In fact, in our experiments this often happens, and it is mainly due to the inter-classifier competition constraint. This is apparently in contrast with the common deep-learning training practice, where the trend is to use as much data as possible. However, in a WSD scenario, adding data which are wrong (noisy) most likely does not improve the training quality. For instance, Fig.~\ref{intro:qualitative_results}, first column, shows the {\em class-specific} top score box $z_y$ for a set of random image samples, computed taking into account the corresponding image label ($y$) and without inter-classifier competition. In a typical MIL-like approach $z_y$ is selected as pseudo-ground truth, and in most of the cases these data are very noisy. Conversely, our self-paced learning strategy leads to select (and use) $z_y$ only in later stages, most of the times decreasing the overall amount of noise (see Fig.~\ref{intro:qualitative_results}, columns 1-4). 

It is also worth noticing that our choice of using $T_t$  for only one epoch of SGD-training is important to avoid overfitting, especially when $N_t = |T_t|$ is initially small (see  the {\em  Computational issues} paragraph).

Finally, the sequence of datasets $T_1, ... T_M$ is not monotonic, meaning that an image $I \in T_{t_1}$ can be discarded when $T_{t_2}$ is created (with $t_1 < t_2$). Empirically we observed frequent oscillations in the class-specific Average Precision (AP) when training both on Pascal VOC 07 and on 
ILSVRC 2013. However, the overall mean AP is almost monotonic with respect to the self-paced iterations, with some small oscillations (typically less than $1\%$), showing that the combined effect of   weight sharing among classes 
and  the common background class (whose samples are included in each mini-batch, independently of the positive class $y$) leads the  network as a whole to benefit from the progressive increase of training data through time.
In Sec.~\ref{Ablation} we analyse this progressive behaviour of our networks and we contrast it with respect to other MIL-like solutions.

\section{Detection performance}
\label{WeaklySupervisedObjctDetection}

In this section we compare our method with other WSD approaches using the two most common WSD datasets (Pascal VOC  and ILSVRC 2013). Comparing  to each other different  methods developed in the past years is not easy due to their heterogeneity, which has been increased  after the introduction of 
deep-learning approaches. For instance, some methods \cite{jacob} use SVMs and hand-crafted features, others  \cite{DBLP:journals/pami/CinbisVS17,Song1,Song2,Bilen1,DBLP:conf/cvpr/BilenPT15,DBLP:journals/tip/WangHRZM15} use SVMs and pre-trained CNN-features obtained with the AlexNet  and very recently a few  deep learning approaches based on an end-to-end training \cite{Bilen16,kantorov2016,Li_2016_CVPR,attentionBMVC2016,DBLP:conf/eccv/ShiF16} have been introduced (including ours). Moreover,   
the latter usually present results obtained with small (e.g. AlexNet \cite{DBLP:conf/nips/KrizhevskySH12}), medium (e.g. VGG-CNN-M-1024 from \cite{DBLP:conf/bmvc/ChatfieldSVZ14}) or large (e.g. VGG-16 \cite{DBLP:journals/corr/SimonyanZ14a}) capacity networks or using an ensemble of three or more architectures.

In order to have a comparison which is the most fair as possible, we 
 separate methods based on low-capacity, AlexNet-like networks from those based on higher-capacity, VGG-16-like architectures and we  mark those approaches which use ensembles of networks. In the rest of the paper, if not otherwise explicitly specified, the basic architecture used for our experiments (initialization included) is AlexNet.

{\bf The ILSVRC 2013 detection dataset} \cite{DBLP:journals/corr/RussakovskyDSKSMHKKBBF14} is a standard benchmark for object detection.
It is partitioned in 3 main subsets: {\em train}, {\em val} and {\em test}.
The {\em train} images are more object-centric (one or very few objects per image on average) and represent more classification-style data than the images in the other 2 partitions 
\cite{DBLP:conf/cvpr/HoffmanPDS15,DBLP:journals/corr/HoffmanGTDGDS14,DBLP:conf/cvpr/GirshickDDM14}. 
All the images of the ILSVRC 2013 detection dataset are annotated with object-level ground truth that we do {\em not} use. 
We   use {\em only} the labels of the objects contained in each image (where each label ranges over $C = 200$ classes).
Girshick et al. \cite{DBLP:conf/cvpr/GirshickDDM14}  split {\em val} further in {\em val1} and {\em val2} and use at most 1000 randomly selected images per category from {\em train}. We use the same approach and $\approx 200 K$ randomly selected images from {\em train} were used to fine-tune AlexNet and obtain $h^I$ as explained in Sec.~\ref{WeaklySelf-Paced} and 
following the procedure suggested in
 \cite{DBLP:journals/corr/HoffmanGTDGDS14,DBLP:conf/cvpr/HoffmanPDS15}. Then, we use {\em only val1} as our set $T$ ($N \approx 20 K$, after image mirroring). $T$ is the training dataset used both in $Init$ and in Alg.~\ref{alg.SelfPaced.training}.
Finally, we evaluate on {\em val2}, whose cardinality is $\approx 10 K$ (the test images are not mirrored). 
Note that this is a broadly adopted protocol, both for supervised (e.g.,\cite{DBLP:conf/cvpr/GirshickDDM14,DBLP:conf/iccv/Girshick15}) and weakly or semi-supervised (e.g., \cite{Li_2016_CVPR,DBLP:journals/corr/HoffmanGTDGDS14,DBLP:conf/cvpr/HoffmanPDS15}) object detection experiments.

Once our network is trained using $Init$ and Alg.~\ref{alg.SelfPaced.training},  it is used as a standard detector at testing time. In other words, given a test image $I$, we apply Non-Maxima Suppression on $f(I, B(I))$ as in the original Fast-RCNN approach \cite{DBLP:conf/iccv/Girshick15} and in this way we obtain {\em multiple, spatially separated detections} per category on $I$ and we can compute Average Precision (AP) and mean Average Precision (mAP) following the standard object detection protocol \cite{pascal-voc-2007}.

In Tab.~\ref{tab.Imagenet}  we compare our approach with the previously published  WSD results on ILSVRC 2013:
Wang et al. \cite{DBLP:journals/tip/WangHRZM15} and Li et al. \cite{Li_2016_CVPR}.
Our method largely outperforms the state of the art in this large dataset. Note that 
Li et al. \cite{Li_2016_CVPR} report other results obtained using a VGG network \cite{DBLP:journals/corr/SimonyanZ14a}: 10.8 mAP, which is  lower than what we obtained with our self-paced training protocol using a much smaller network.

\begin{table}[h]
\centering
\small
\begin{tabular}{|c|c|c|}
\hline
Method	& End-to-end training   & mAP		\\ \hline 
Wang et al.$^{(*)}$  \cite{DBLP:journals/tip/WangHRZM15}	&   		&   6    \\ \hline
Li et al.  \cite{Li_2016_CVPR}	&   $\surd$		&   7.7   \\ \hline
Ours	&    $\surd$		&   {\bf 12.13}    \\ \hline
\end{tabular}
\caption{Quantitative comparison (mAP $\%$) on the ILSVRC 2013  detection 
dataset. All the methods are based on a single, low-capacity network (AlexNet).  Wang et al. \cite{DBLP:journals/tip/WangHRZM15} use AlexNet to extract  CNN-features from the BBs but do not perform end-to-end training. $(*)$ Note that the result of Wang et al. \cite{DBLP:journals/tip/WangHRZM15} is obtained on the whole ILSVRC 2013 {\em val} set, while our result and the result of Li et al. \cite{Li_2016_CVPR} are computed on the {\em val2} split.}
\label{tab.Imagenet}
\end{table}


 
\begin{table*}[!htbp] 
\footnotesize
\centering
\addtolength{\tabcolsep}{-5pt}    
\hspace*{-1cm}{
\begin{tabular}{|c|c|cccccccccccccccccccc|c|}
\hline
Method	& End-to-end    & aero & bike & bird & boat & bottle & bus & car & cat & chair & cow & table & dog & horse & mbike & persn & plant & sheep & sofa & train & tv & mAP		\\ \hline 
Song et al.  \cite{Song1}	&   		& 27.6  & 41.9 & 19.7 & 9.1 & 10.4 & 35.8 & 39.1 & 33.6 & 0.6 & 20.9 & 10 & 27.7 & 29.4 & 39.2 & 9.1 & 19.3 & 20.5 & 17.1 & 35.6 & 7.1 & 22.7    \\ \hline
Song et al. \cite{Song2}	&   		& 36.3 & 47.6 & 23.3 & 12.3 & 11.1 & 36 & 46.6 & 25.4 & 0.7 &  23.5 & 12.5 & 23.5 & 27.9 & 40.9 & 14.8 & 19.2 & 24.2 & 17.1 & 37.7 & 11.6  & 24.6   \\ \hline
Bilen et al. \cite{Bilen1}	&   		& 42.2 & 43.9 & 23.1 & 9.2 & 12.5 & 44.9 & 45.1 & 24.9 & 8.3 & 24 & 13.9 & 18.6 & 31.6 & 43.6 & 7.6 & {\bf 20.9} & 26.6 & 20.6 & 35.9 & 29.6  & 26.4   \\ \hline
Bilen et al. \cite{DBLP:conf/cvpr/BilenPT15}	&   	& 46.2 & 46.9 & 24.1 & 16.4 & 12.2 & 42.2 & 47.1 & 35.2 & 7.8 & 28.3 & 12.7 & 21.5  & 30.1 & 42.4 & 7.8 & 20 & 26.8 & 20.8 & 35.8 & 29.6 & 27.7   \\ \hline
Cinbis et al. \cite{DBLP:journals/pami/CinbisVS17}	&   		& 39.3 & 43 & 28.8  & 20.4 & 8 & 45.5 & 47.9 & 22.1 & 8.4 & 33.5 & 23.6 & 29.2 & 38.5 & 47.9 & 20.3 & 20 & 35.8 & 30.8 & 41 & 20.1  & 30.2   \\ \hline
Wang et al.$^{(*)}$ \cite{DBLP:journals/tip/WangHRZM15}	&   	& 48.9  & 42.3  & 26.1 & 11.3 & 11.9 & 41.3 & 40.9 & 34.7 & 10.8 & 34.7 & 18.8 & 34.4 & 35.4 & 52.7 & 19.1 & 17.4 & 35.9 & 33.3 & 34.8 & 46.5 & 31.6   \\ \hline
Li et al. \cite{Li_2016_CVPR}	&    $\surd$		& 49.7 & 33.6 & 30.8 & 19.9 & 13 & 40.5 & 54.3 & 37.4 & 14.8 & 39.8 & 9.4 & 28.8 & 38.1  & 49.8 & 14.5 & 24 & 27.1 & 12.1 & 42.3 & 39.7  & 31.0    \\ \hline
Bilen et al. \cite{Bilen16}	&    $\surd$	 & 42.9 & {\bf 56} & 32 & 17.6 & 10.2 & {\bf 61.8} & 50.2 & 29 & 3.8  & 36.2 & 18.5 & 31.1 & 45.8 & 54.5 & 10.2 & 15.4 & 36.3 &  45.2 & 50.1	& 43.8  & 34.5    \\ \hline
Shi et al.$^{(**)}$ \cite{DBLP:conf/eccv/ShiF16}	&    $\surd$		&  - & - & - & - & - & - & - & - & - & - & - & - & - & - & - & - & - & - & - & - & 36.0    \\ \hline
Kantorov et al. \cite{kantorov2016}	&    $\surd$		&  {\bf 57.1} & 52 & 31.5 & 7.6 & 11.5 & 55 & 53.1 & 34.1 & 1.7 & 33.1 & {\bf 49.2} & {\bf 42} & {\bf 47.3} & 56.6 & 15.3 & 12.8 & 24.8 & {\bf 48.9} & 44.4 & {\bf 47.8} & 36.3    \\ \hline
Ours ($W_0^{I,P}$)	&    $\surd$		& 49.2 & 43.7 & 34.9 & {\bf 26.11} & 3.9 & 47.8 & 53.3 & 48.4 & 13.8 & 14.7 & 0.7 & 28.1 & 36.6 & 49.3 & {\bf 21.7} & 16.7 & 26.8 & 31.9 & 52.5 & 42 & 32.1   \\ \hline
Ours ($W_0^P$)	&    $\surd$		&  50.1 & 52.4 & {\bf 35.8} & 22.9 & {\bf 13.4} & 55.3 & {\bf 56.3} & {\bf 58.9} & {\bf 17.7} & {\bf 46.2} &  30.1 &  40.3 & 44.6 & {\bf 57.5} & 8.6 & 16.2 & {\bf 39.7} & 24.5  & {\bf 54.6} & 37.1 &  {\bf 38.11}    \\ \hline
\end{tabular}}\hspace*{-1cm}
\caption{Quantitative comparison (AP $\%$) on the Pascal VOC 2007  {\em test} 
set.  All the methods are based on a single network with a capacity comparable with
AlexNet, which is used either as a black-box to  extract CNN-features or to perform end-to-end training (see the corresponding column). Note that: $(*)$ Wang et al. \cite{DBLP:journals/tip/WangHRZM15} 
specifically 
tune  the number of latent categories for each class  and  $(**)$ Shi et al. \cite{DBLP:conf/eccv/ShiF16} use additional data (the Pascal VOC 2012 dataset) with BB-level annotation at training time. } 
\label{tab.Pascal-mAP}
\end{table*} 


\begin{table*}[!htbp]
\footnotesize
\centering
\addtolength{\tabcolsep}{-5pt}    
\hspace*{-1cm}{
\begin{tabular}{|c|c|cccccccccccccccccccc|c|}
\hline
Method	& End-to-end   & aero & bike & bird & boat & bottle & bus & car & cat & chair & cow & table & dog & horse & mbike & persn & plant & sheep & sofa & train & tv & avg		\\ \hline 
Bilen et al. \cite{DBLP:conf/cvpr/BilenPT15}	&   		& 66.4 & 59.3 & 42.7 & 20.4 & 21.3 & 63.4 & 74.3 & 59.6 & 21.1 & 58.2 & 14 & 38.5 & 49.5 & 60 & 19.8 & 39.2 & 41.7 & 30.1 & 50.2 & 44.1  & 43.7   \\ \hline
Cinbis et al. \cite{DBLP:journals/pami/CinbisVS17}	&     & 65.3  & 55 & 52.4 & {\bf 48.3} & 18.2 & 66.4 & {\bf 77.8} & 35.6 & 26.5 & 67 & 46.9 & 48.4 & {\bf 70.5} & 69.1 & 35.2 & 35.2 & 69.6 & 43.4 & {\bf 64.6} & 43.7  & 52.0   \\ \hline
Wang et al.$^{(*)}$ \cite{DBLP:journals/tip/WangHRZM15}	&   		&  80.1 & 63.9 & 51.5 & 14.9 & 21 & 55.7 & 74.2 & 43.5 & 26.2 & 53.4 & 16.3 & 56.7 & 58.3 & 69.5 & 14.1 & 38.3 & 58.8 & 47.2 & 49.1 & 60.9 &  48.5   \\ \hline
Li et al. \cite{Li_2016_CVPR}	&    $\surd$		& 77.3 & 62.6 & 53.3 & 41.4 & 28.7 & 58.6 & 76.2 & 61.1 & 24.5 & 59.6 & 18 & 49.9 & 56.8 & 71.4 & 20.9 & {\bf 44.5} & 59.4 & 22.3 & 60.9 & 48.8 &  49.8    \\ \hline
Bilen et al. \cite{Bilen16}	&    $\surd$		& 68.5 & 67.5 & {\bf 56.7} & 34.3 & 32.8 & 69.9 & 75 & 45.7 & 17.1 & {\bf 68.1} & 30.5 & 40.6 & 67.2 & 82.9 & 28.8 & 43.7 & {\bf 71.9} & {\bf 62} & 62.8 & 58.2  & 54.2    \\ \hline
Shi et al.$^{(**)}$ \cite{DBLP:conf/eccv/ShiF16}	&    $\surd$		& - & - & - & - & - & - & - & - & - & - & - & - & - & - & - & - & - & - & - & -  & {\bf 60.9}    \\ \hline
Kantorov et al. \cite{kantorov2016}	&    $\surd$		&  {\bf 83.3} & 68.6 & 54.7 & 23.4 & 18.3 & {\bf 73.6} & 74.1 & 54.1 & 8.6 & 65.1 & {\bf 47.1} & 59.5 & 67 & {\bf 83.5} & {\bf 35.3} & 39.9 & 67 & 49.7 & 63.5 & {\bf 65.2} & 55.1    \\ \hline
Ours ($W_0^{I,P}$)	&    $\surd$		 & 68.6 & 55.2 & 49 & 47 & 18 & 68.3 & 73.5 & 70.6 & 22.1 & 23.3 & 1.9 & 40.3 & 61 & 69.4 & 31.7 & 37.7 & 38.9 & 44.3 & 59.4 & 60.5  &  47.0  \\ \hline
Ours ($W_0^P$)	&    $\surd$	&  76.4 & {\bf 70.3} & 54.8 & 40.7 & {\bf 36.2} & 72 & 73.4 & {\bf 76.8} & {\bf 46} & 58.9 & 43.9 & {\bf 61.3} & 67.4 & 75 & 27 & 38.4 & 60.4 & 55.9 & 62.8 & 61.9 & 58.0    \\ \hline
\end{tabular}}\hspace*{-1cm}
\caption{Quantitative comparison (CorLoc $\%$) on the Pascal VOC 2007  {\em trainval} 
set (with a single AlexNet-like capacity network). Note that:
$(*)$ Wang et al. \cite{DBLP:journals/tip/WangHRZM15} 
tune  the number of latent categories for each class  and 
 $(**)$ Shi et al. \cite{DBLP:conf/eccv/ShiF16} use additional data  with BB-level annotation at training time.}
\label{tab.Pascal-CorLoc}
\end{table*}


{\bf Pascal VOC 2007} 
\cite{pascal-voc-2007}  
is another well known benchmark for object detection and it is widely used by different WSD methods 
\cite{Bilen16,kantorov2016,Li_2016_CVPR,attentionBMVC2016,DBLP:conf/eccv/ShiF16,DBLP:journals/pami/CinbisVS17,jacob,Song1,Song2,Bilen1,DBLP:conf/cvpr/BilenPT15,DBLP:journals/tip/WangHRZM15}. The number of classes ($C$) is 20. We adopted the common training-testing protocol  in which training is done on the {\em trainval} split  and testing is performed on the {\em test} split.
Hence, $T$ is {\em trainval} in which {\em only image-level labels were used}. Since {\em trainval} is highly imbalanced toward the $person$ class, we have subsampled this class, selecting the 262 top-score images used in case of $Init$.  
Since we used different CNs to initialize the weights of our network (see Sec.~\ref{WeaklySelf-Paced}), 
we report results obtained using $W_0 = W_0^{I,P}$ and $W_0 = W_0^P$ in Alg.~\ref{alg.SelfPaced.training}, respectively.
The results are shown in Tab.~\ref{tab.Pascal-mAP}, where all the methods (including ours) use AlexNet (or an AlexNet-like) as the basic architecture.
 Using a CN trained on ILSVRC
($h^I$, see Sec.~\ref{WeaklySelf-Paced}), the final mAP reached after the self-paced training (Alg.~\ref{alg.SelfPaced.training}) is 32.1, which is comparable with other state-of-the-art WSD methods. Using a stronger initialization, in which the CN is trained using a multi-label loss \cite{Li_2016_CVPR} directly on Pascal VOC ($h^P$,   Sec.~\ref{WeaklySelf-Paced}), the final mAP  after the self-paced training is 38.11, which is higher than all the other weakly supervised approaches tested on this dataset. 
  
  In order to show how large is the gain that can be obtained using the proposed self-paced training protocol, we compare the detection performance achieved when using only $Init$ for fine-tuning the Fast-RCNN (see Sec.~\ref{WeaklySelf-Paced}) with the performance obtained after the self-paced phase ({\em SP}) described in 
Alg.~\ref{alg.SelfPaced.training}.
Since $Init$ is used to initialize Alg.~\ref{alg.SelfPaced.training}, the accuracy difference  shows the boost  obtained by the self-paced strategy.
The results,  reported in Tab.~\ref{tab.Pascal-initialization-gain},  
 show that, independently of the CN used to initialize $Init$, in both cases the {\em SP}-based  boost over the $Init$ phase is dramatic: $+7.36$ mAP in case of $h^I$ and $+6.18$ in case of $h^P$.
Moreover, since we trained   $h^P$ as proposed by Li et al. \cite{Li_2016_CVPR}, it is also interesting to directly compare the results we obtained using $h^P$  and the results obtained in \cite{Li_2016_CVPR}, where a {\em detection} network is trained on top of the same 
CN. Our ``simple'' $Init$-based detector achieves a slightly better mAP (31.93) than  
the mAP (31) achieved by the DN proposed in  \cite{Li_2016_CVPR}. However, our full-pipeline $Init + SP$ obtains a much higher mAP value (38.11), showing the advantage of a self-paced training strategy.
The gap with respect to \cite{Li_2016_CVPR} is even higher in other experiments (e.g., see Tabs.~\ref{tab.Pascal10-AlexNet}-\ref{tab.Pascal10-VGG16}).
In Sec.~\ref{Ablation} (Tab.~\ref{experiments:tab2}) we show the corresponding relative boost obtained by {\em SP} with respect to $Init$ 
in case of the ILSVRC dataset.

\begin{table}[h]
\centering
\begin{tabular}{|c|c|c|}
\hline
          		& \multicolumn{2}{c|}{Classification Network}	  	\\ \hline 
          		& $h^I$ 		& $h^P$   	\\ \hline
  $Init$	  	&    24.74		&   31.93    	\\ \hline
  $Init + SP$ 	  		& 	 32.1		&   38.11      	\\ \hline
\end{tabular}
\caption{mAP ($\%$) on Pascal VOC 2007 {\em test} computed using only $Init$ (first row) or $Init + SP$ (second row) for fine-tuning the detector. The first column corresponds to the results obtained when  $Init$ is initialized with a CN trained with only  ILSVRC data ($h^I$), while the second column corresponds to using a CN trained with Pascal VOC data and a multi-label loss \cite{Li_2016_CVPR} ($h^P$).}
\label{tab.Pascal-initialization-gain}
\end{table}

In Tab.~\ref{tab.Pascal-CorLoc} we report the CorLoc computed on the {\em trainval} split, which is a common metric adopted in many WSD approaches tested on Pascal VOC 07.
Our CorLoc result corresponding to the $W_0^P$-based initialization is the second best after the values obtained by Shi et al. \cite{DBLP:conf/eccv/ShiF16}.

Finally,  in Tab.~\ref{tab.Pascal07-VGG} we report the results obtained using a larger capacity network. Specifically, we use VGG-16 \cite{DBLP:journals/corr/SimonyanZ14a} and we compare with those works which use the same (single) network.
In our case, VGG-16 is used both at initialization time, where we train a  CN ($h^P$) following the steps reported in Sec.~\ref{WeaklySelf-Paced}, and for the Fast-RCNN DN, trained using  our self-paced protocol ($Init + SP$). As shown in Tab.~\ref{tab.Pascal07-VGG}, we obtain state-of-the-art results, largely outperforming the other methods. The importance of this experiments relies on the fact that it demonstrates that our self-paced  protocol is not inclined to overfitting with a large-capacity network.


\begin{table*}[!htbp] 
\footnotesize
\centering
\addtolength{\tabcolsep}{-5pt}    
\hspace*{-1cm}{
\begin{tabular}{|c|cccccccccccccccccccc|c|}
\hline
Method	   & aero & bike & bird & boat & bottle & bus & car & cat & chair & cow & table & dog & horse & mbike & persn & plant & sheep & sofa & train & tv & mAP		\\ \hline 
Li et al. \cite{Li_2016_CVPR}			& 54.5 & 47.4 & 41.3 & 20.8 & 17.7 & 51.9 & 63.5 & 46.1 & 21.8 & 57.1 & 22.1 & 34.4 & 50.5  & 61.8 & {\bf 16.2} & {\bf 29.9} & 40.7 & 15.9 & 55.3 & 40.2  & 39.5    \\ \hline
Bilen et al. \cite{Bilen16}		 & 39.4 & 50.1 & 31.5 & 16.3 & 12.6 & {\bf 64.5} & 42.8 & 42.6 & 10.1  & 35.7 & 24.9 & {\bf 38.2} & 34.4 & 55.6 & 9.4 & 14.7 & 30.2 & {\bf 40.7} & 54.7	& 46.9  & 34.8    \\ \hline
Shi et al.$^{(*)}$ \cite{DBLP:conf/eccv/ShiF16}			&  - & - & - & - & - & - & - & - & - & - & - & - & - & - & - & - & - & - & - & - & 37.2    \\ \hline
Ours ($W_0^P$)			& {\bf 61.7} & {\bf 66.7} & {\bf 47.7} & {\bf 26.2} & {\bf 19.1} & 61.3 & {\bf 68} & {\bf 52.6} & {\bf 23.5} & {\bf 61.8} & {\bf 26.5} &  27.8 & {\bf 57.3} & {\bf 63.7} & 14.3 & 24.6 & {\bf 46.3} & 31.2  & {\bf 66.8} & {\bf 49.5} &  {\bf 44.84}    \\ \hline
\end{tabular}}\hspace*{-1cm}
\caption{Quantitative comparison (AP $\%$) on the Pascal VOC 2007  {\em test} 
set using a single VGG-16 network.  Note that $(*)$ Shi et al. \cite{DBLP:conf/eccv/ShiF16} use additional data  with BB-level annotation at training time.} 
\label{tab.Pascal07-VGG}
\end{table*} 


{\bf Pascal VOC 2010} \cite{pascal-voc-2010}. In Tabs.~\ref{tab.Pascal10-AlexNet}-\ref{tab.Pascal10-VGG16} we show the results obtained with the Pascal VOC 2010 {\em test} set.  Tab.~\ref{tab.Pascal10-AlexNet} refers to methods using an AlexNet-like network, while Tab.~\ref{tab.Pascal10-VGG16} refers to methods using a VGG-16 or an ensemble of VGG-16 and other networks. In both cases, our results are the state of the art. Specifically, our VGG-16 based results are even higher than previous results  obtained with an ensemble of 3 networks \cite{Bilen16}.



\begin{table*}[!htbp] 
\footnotesize
\centering
\addtolength{\tabcolsep}{-5pt}    
\hspace*{-1cm}{
\begin{tabular}{|c|cccccccccccccccccccc|c|}
\hline
Method	   & aero & bike & bird & boat & bottle & bus & car & cat & chair & cow & table & dog & horse & mbike & persn & plant & sheep & sofa & train & tv & mAP		\\ \hline 
Cinbis et al. \cite{DBLP:journals/pami/CinbisVS17}	&      44.6  & 42.3 & 25.5 & 14.1 & 11 & 44.1 & 36.3 & 23.2 & {\bf 12.2} & 26.1 & {\bf 14} & 29.2 & 36 & 54.3 & {\bf 20.7} & 12.4 & 26.5 & {\bf 20.3} & 31.2 & 23.7 & 27.4   \\ \hline
Li et al.$^{(*)}$ \cite{Li_2016_CVPR}			&  - &  - &  - &  - &  - &  - &  - &  - &  - &  - &  - &  - &  -  &  - &  - &  - &  - &  - &  - &  -  & 21.4    \\ \hline
Ours ($W_0^P$)			& {\bf 55.8} & {\bf 47.5} & {\bf 36.1} & {\bf 16} & {\bf 12.5} & {\bf 46.3} & {\bf 43.7} & {\bf 59.3} &  11.9 & {\bf 32.2} &  13.4 &  {\bf 41.1} & {\bf 40.3} & {\bf 55.8} & 8 & {\bf 13} & {\bf 33.8} & 15.6  & {\bf 42.3} & {\bf 28.5} &  {\bf 32.66}    \\ \hline
\end{tabular}}\hspace*{-1cm}
\caption{Pascal VOC 2010  {\em test} 
set, single AlexNet network.  Note that $(*)$ Li et al. \cite{Li_2016_CVPR} use the {\em val} split for the evaluation. Our results are available at: \url{http://host.robots.ox.ac.uk:8080/anonymous/WHEGLB.html}.} 
\label{tab.Pascal10-AlexNet}
\end{table*}


\begin{table*}[!htbp] 
\footnotesize
\centering
\addtolength{\tabcolsep}{-5pt}    
\hspace*{-1cm}{
\begin{tabular}{|c|cccccccccccccccccccc|c|}
\hline
Method	   & aero & bike & bird & boat & bottle & bus & car & cat & chair & cow & table & dog & horse & mbike & persn & plant & sheep & sofa & train & tv & mAP		\\ \hline 
Li et al.$^{(*)}$ \cite{Li_2016_CVPR}			&  - &  - &  - &  - &  - &  - &  - &  - &  - &  - &  - &  - &  -  &  - &  - &  - &  - &  - &  - &  -  & 30.7    \\ \hline
Bilen et al.$^{(**)}$ \cite{Bilen16}		 & 57.4 & 51.8 & 41.2 & 16.4 & {\bf 22.8} & {\bf 57.3} & 41.8 & 34.8 & 13.1  & 37.6 & 10.8 & {\bf 37} & 45.2 & 64.9 & {\bf 14.1} & {\bf 22.3} & 33.8 & {\bf 27.6} & 49.1	& {\bf 44.8}  & 36.2    \\ \hline
Ours ($W_0^P$)			& {\bf 70.4} & {\bf 60.7} & {\bf 45.7} & {\bf 26} & 15.6 & 54.8 & {\bf 54.5} & {\bf 53.7} & {\bf 18.6} & {\bf 42.8} & {\bf 11.2} &  35.6 & {\bf 53} & {\bf 67} & 6 & 21.8 & {\bf 38.3} & 24.8  & {\bf 60.8} &  44.2 &  {\bf 40.27}    \\ \hline
\end{tabular}}\hspace*{-1cm}
\caption{Pascal VOC 2010  {\em test} 
set, VGG-16 network.  Note that: $(*)$ Li et al. \cite{Li_2016_CVPR} use the {\em val} split for the evaluation and $(**)$ Bilen et al. \cite{Bilen16} use an ensemble of 3 networks: VGG-F \cite{DBLP:conf/bmvc/ChatfieldSVZ14}, VGG-CNN-M-1024 \cite{DBLP:conf/bmvc/ChatfieldSVZ14} and  VGG-16. Our results are available at: \url{http://host.robots.ox.ac.uk:8080/anonymous/UBLBGP.html}.} 
\label{tab.Pascal10-VGG16}
\end{table*} 


\section{Analysis of different aspects of the protocol}
\label{Analysis}

In this section we  analyse the influence  of different elements of our proposed training protocol by separately removing or modifying important parts of Alg.~\ref{alg.SelfPaced.training}.

\subsection{Simplified versions of the training protocol}
\label{Ablation}

{\bf Basic-MIL.}
In the experiments of this subsection we 
use both Pascal VOC 07 and ILSVRC 2013.
We start with comparing our method (Self-Paced, {\em SP}) with a 
 MIL-based solution ({\em MIL}), where: (a) all the images in $T$ are used and (b) in each image the latent boxes are computed by iteratively maximizing the {\em class-specific} score of the current iteration's model. 
 Thus, we remove from Alg.~\ref{alg.SelfPaced.training} all those steps which concern image (and class) selection. Moreover, we also remove the inter-classifier competition, and we independently select the top score box {\em for each label} in $Y$.
More in detail, given $(I,Y) \in T$, {\em for each} $y \in Y$ we separately compute:

\begin{equation}
  (s_y, z_y)  = \argmax_{ \substack{(s_{ic}, p_{ic}) \in f_{W_{t-1}}(I, B(I)),  \\ c = y}} s_{ic}    
  \label{latent-box-mil} 
\end{equation}

\noindent
Note that Eq.~\ref{latent-box-mil} is different from Eq.~\ref{latent-box} because only $y$-specific scores are taken into account. 
For instance, an image $I$ associated with 3 different labels ($Y = \{y_1, y_2, y_3 \}$) will produce 3 corresponding class-specific latent boxes ($\{z_{y_1}, z_{y_2}, z_{y_3} \}$) using Eq.~\ref{latent-box-mil} and iterating over $y \in Y$, but at most only one latent box $z_y^I$ using Eq.~\ref{latent-box} and the inter-classifier competition constraint (see Sec.~\ref{WeaklySelf-Paced}).
In  {\em MIL} we use Eq.~\ref{latent-box-mil} to associate
each $I$  with the set of its pseudo-ground truth boxes $(I, \{ (y, z_y) \}_{y \in Y})$, which are directly added to $T_t$, skipping Lines 4-12. 
Note also that  Eq.~\ref{latent-box-mil} is based on the current network $f_{W_{t-1}}$, including the regression part, which is used at each iteration $t$ to compute the set $\{  p_{ic} \}$ over which $z_y$  is selected. Thus, both in {\em MIL} and in {\em SP} the bag of boxes associated with $I$ dynamically changes at each iteration $t$.
Finally, note that
 {\em MIL} is different from $Init$, since in the latter  the pseudo-ground truth is computed only once and using the CN. However, both in {\em MIL} and in $Init$, $N_t = N = |T|$ is kept fixed in every iteration $t$ and no
score-based image selection or class selection is done, hence we use {\em all} the images for training. 

{\bf Curriculum.}
In contrast, we call {\em Curriculum}  a simplified version of Alg.~\ref{alg.SelfPaced.training} in which information about ``easiness'' of images and boxes is given {\em externally} to the trained DN.
The purpose of this experiment is to show the behaviour of a simplified 
 progressive-image-selection   protocol
in which image selection is performed ``statically'' using the scores computed by the CN, as opposed to  {\em SP}, where  selection is performed ``dynamically'' using the {\em current} DN $f_{W_{t-1}}$.
In {\em Curriculum} we use the same pseudo-ground truth used in  $Init$ and we  select images according to the static score values  initially computed by the CN.
 Specifically, we use the boxes and the corresponding scores computed when we build the the training set of $Init$  (Sec.~\ref{WeaklySelf-Paced}):
 For each $(I,Y) \in T$ and {\em for each} $y \in Y$, $(s_y, z_y)$ is obtained using 
 Eq.~\ref{MIL1} and the CN ($h^I$ and $h^P$ for ILSVRC and Pascal, respectively). Thus, $T$ is the same used in $Init$. However, at each iteration $t$, $T$ is sorted using the pre-computed scores ($s_y$) and a subset $T_t$ is extracted from $T$ as in Lines 10-11. 
 When we select $T_t$ we use  the same  ratio sequence $r_1, ... r_t, ... r_M$ used in {\em SP} (see Alg.~\ref{alg.SelfPaced.training}).
However,  
no inter-classifier competition and thus no class-selection is used in {\em Curriculum}.
 Images associated with multiple labels are sorted according to their highest score label.
Note that in this experiment, despite the values of the latent boxes and their scores are fixed, the model will observe more and more data (from 
 ``easy'' to ``difficult'' images) 
 while $t$ increases.

The results are shown in Tab.~\ref{experiments:abaltion-pascal} and \ref{experiments:tab2} for the two datasets, where we report the mAP 
for different networks $f_{W_t}$ obtained at the end of different iterations $t$ and for each of the training protocols. 
For all the training procedures $M = 4$ and $r_{t+1}$ is computed as in Line 19 
(when it applies).
However,
in the last column of the table we show the results obtained iterating  all the protocols for one more iteration (using $r_{M+1} = 1$). The accuracy impact of the last iteration  is generally marginal. 
The 
first column of the table ($W_0$) is the same for all the  methods as it is the evaluation of $Init$, 
which is used as the pre-trained model for all the protocols.
The results  show that {\em SP} is able to  increase its accuracy during time, confirming the self-paced assumption of a model which progressively becomes more mature and, as a consequence, is more and more reliable when it computes  the values of its latent variables.

On the other hand, the other two simplified solutions do not really seem to be able to improve over time.
In both datasets {\em MIL} achieves a final mAP even {\em worse} than $Init$ (both when $t = M$ and when $t = M + 1$).
This is probably due to a drifting effect: At time $t$  {\em all} the images in $T$ are used for training and a large portion  of them are noisy (i.e., are associated with wrong latent boxes). Thus $W_t$ is a weak model and it is used to compute $T_{t+1}$, most likely producing many errors that will be accumulated over time.
In {\em SP} this problem is alleviated because only the ``best'' images in $T_t$ are used to train
$W_t$.
We do not claim that {\em any} MIL-based solution should drift, and, for example, Cinbis et al. \cite{DBLP:journals/pami/CinbisVS17} combined a standard MIL approach with a multi-fold strategy, observing a progressively increasing accuracy in their experiments. However, the baseline we adopted in our experiments, which is based on the ``standard'' MIL characteristics (see points (a) and (b) above), embedded in our framework, did not show significant progresses in the two datasets we used for the evaluation, confirming the difficulty  of making MIL-like methods work in practice. 

On the other hand,
sample selection is performed  in
{\em Curriculum}, but, similarly to {\em MIL}, the mAP values oscillate without improving with respect to 
the starting point.
This is probably due to the fact that there can be little progress if the model cannot update the initial predictions done by the CN. 
Moreover, the sample selection strategy in {\em SP} takes also into account the inter-classifier competition,
 prudently discarding those images in which the current model is uncertain, a constraint which is not used in {\em Curriculum}. 
 In  Sec.~\ref{Precision} we present  another  experiment showing the precision of the  latent boxes computed by {\em SP},  which further confirms this hypothesis. 

{\bf Class selection.}
In the same tables we show the results of {\em SP-all-cls}, obtained from {\em SP} by removing {\em only} the class-selection part (Lines 6-9) 
and keeping all the rest unchanged (inter-classifier competition included).
In both the datasets {\em SP-all-cls} achieved a lower final mAP with respect to {\em SP}, but this is more evident in the case of Pascal.
In the latter dataset we performed an additional experiment, called {\em SP-rnd-cls}, where class selection is performed {\em randomly}, in such a way that the number of classes 
($C_t$, see Line 7 
of Alg.~\ref{alg.SelfPaced.training}) is the same used in {\em SP} and the results are reported in Tab.~\ref{experiments:abaltion-pascal}.
{\em SP-rnd-cls} is able to improve   with respect to the initialization but, as expected, the final mAP is much lower than the corresponding mAP obtained with {\em SP-all-cls}.
The reason of this behaviour is 
 most likely due to the fact that, randomly discarding easy classes and the corresponding correct samples, has a negative effect on the final performance.
Comparing the category-specific AP obtained with $Init$ in ILSVRC with the corresponding AP obtained by {\em SP} (see Sec.~\ref{Appendix}), 147 classes out of  200  ($73.5\%$) have improved. 
This demonstrates that the combination of class-selection and inter-classifier competition in {\em SP} does {\em not} prevent most of the classes to improve  and that the learning process is {\em not} dominated by few strong classifiers.
Finally, we list below the 10 classes selected when $t=1$  in Pascal  
in descending order: 
$S = \{ aeroplane, sheep, train, cow, bird, cat, boat, horse, 
\\motorbike, car \}$.
 As expected, this set includes those ``classifiers'' which are already strong enough both in $h^P$ and in $f_{W_0^P}$, 
and it is largely independent of the cardinality of corresponding class samples in $T$.
 
{\bf Regression.} 
 The last part of this ablation study is dedicated to  the importance of the regression part.
In {\em SP},  $z_y^I$ is selected over the set $\{ p_{ic} \}$,
computed using the regression layer of Fast-RCNN. 
Choosing $z_y^I$ over the BBs in $B(I)$ at training time but using 
 the regression layer at testing time, we obtain the results reported in 
 Tab.~\ref{experiments:abaltion-pascal} ({\em No-reg-train}). Note that we can use the regression layer at testing time because the weights of this layer are trained using the pseudo-ground truth, though that layer is not used for selecting $z_y^I$. Conversely, in {\em No-reg-train-test} we disable the regression layer also at testing time (in this case $f_{W_0}$ does {\em not} correspond to $Init$, being the results of $Init$ obtained using the regression part). 
  As shown in Tab.~\ref{experiments:abaltion-pascal} there is a large accuracy gap with respect to {\em SP}. We repeated the {\em No-reg-train-test} experiment on  ILSVRC obtaining a similar very large accuracy gap: 
 a final mAP of 7.57, 
4.56 points less than our best result on that dataset. These accuracy differences show the importance of the proposed iterative strategy in which the current network is used to compute the supposed locations of the objects inside the training images.



\begin{table}[h]
\centering
\begin{tabular}{|c|c|c|c|c|c|c|}
\hline
    Method      	  	& $W_0$ 	& $W_1$   & $W_2$ 	& $W_3$   & $W_4$   & $W_5$ \\ \hline\hline
  {\em MIL}	  		  	& 31.9  	&  33.6 	&  32.1 	&  32.2 	& 30.8   &  30.9   	\\ \hline
  {\em Curriculum}	  		  	& 31.9  	&  31.3  	&  33.8 	&  31.6 	& 31.3 	 &  30.5   	\\ \hline
  {\em SP-all-cls} 	  	& 31.9  	&  36.6 	&  36.9 	&  36.6 	& 36.9   &  36.9  	\\ \hline
  {\em SP-rnd-cls} 	  	& 31.9  	&  32.3 	&  31.6 	&  32.4 	& 32.7   &  33.8  	\\ \hline
  {\em No-reg-train}  	& 31.9  	&  31.2 	&  32.6 	&  33.1 	& 33.5   &  34.4  	\\ \hline
  {\em No-reg-train-test} & 28.3  	&  28.3 	&  30.1 	&  30.9 	& 30.7   &  31.4  	\\ \hline
  {\em SP} 	  		  	& 31.9  	&  35.3 	&  37.6 	&  37.8 	& 38.1   &  38.1  	\\ \hline
\end{tabular}
\caption{mAP ($\%$) on Pascal VOC 2007 {\em test} computed with different networks $f_{W_t}$ and with respect to different versions of our training protocol and $M + 1$ iterations.}
\label{experiments:abaltion-pascal}
\end{table}


\begin{table}[h]
\centering
\begin{tabular}{|c|c|c|c|c|c|c|}
\hline
    Method      & $W_0$ 	& $W_1$   & $W_2$ 	& $W_3$   & $W_4$   & $W_5$ 	\\ \hline\hline
  {\em MIL}	  & 9.54  &  9.66 &  9.01 &  8.97 & 8.59  &  8.7   	\\ \hline
  {\em Curriculum}	  & 9.54  & 	9.08  &  9.15 &  8.77 &  8.89 & 8.97       	\\ \hline
  {\em SP-all-cls} 	  & 	9.54  &  10.68 &  10.74 &  11.77 & 11.97  &  12.06     	\\ \hline
  {\em SP} 	  & 	9.54  &  10.88 &  11.87 &  12.01 & 12.13  &  11.87     	\\ \hline
\end{tabular}
\caption{mAP ($\%$) on ILSVRC 2013 {\em val2} computed with different networks $f_{W_t}$ and with respect to different versions of our training protocol and $M + 1$ iterations.}
\label{experiments:tab2}
\end{table}


\subsection{Multi-label versions of the training protocol}
\label{SIML-MIML} 
 
 This subsection is dedicated to evaluating the importance of the inter-classifier competition. As explained in Sec.~\ref{WeaklySelf-Paced} the inter-classifier competition  is used in {\em SP} to reduce the amount of noisy training boxes by selecting only one box $z_y^I$ per image $I$, according to the current
most confident classifier ($y$) on  $I$. A disadvantage of this strategy is the 
 sacrifice of possible other ``good'' boxes associated with either other classes ($y' \neq y, y' \in Y$) or the same class. For instance, the latter situation occurs when more than one instance of the category $person$ is contained in $I$ (recall  that we only know that $person \in Y$, without any information about the cardinality of the instances for each image). 
We report below our study of  two different versions of {\em SP}: {\em SP-SIML} and {\em SP-MIML}, where we relax the inter-classifier competition for a multiple-label and a multiple-instance scenario. 
 
{\bf Multiple-label.} 
 {\em SP-SIML} is a Single-Instance-Multiple-Label version of Alg.~\ref{alg.SelfPaced.training},
where in each image we mine  one box {\em per each label} in $Y$. Similarly to {\em MIL}, given $(I,Y) \in T$, {\em for each} $y \in Y$ we use Eq.~\ref{latent-box-mil} to compute a set of candidate boxes associated with $I$: $P_I =  (I, \{ (s_y, z_y, y)\}_{y \in Y } )$.
However, before adding $P_I$ to $P$, in Line 5 
we remove from $P_I$ those boxes which have a score lower than the top-score box of all the other categories $\overline{Y} = \{1, ..., C \} \setminus Y$. $\overline{Y}$ is the complement of $Y$ (it contains the set of all the labels {\em not} included in $Y$) and it is used to compute $s_y^o$:

\begin{equation}
  (s_y^o, z_y^o)  = \argmax_{ \substack{(s_{ic}, p_{ic}) \in f_{W_{t-1}}(I, B(I)),  \\ c \in \overline{Y}}} s_{ic}
  \label{latent-box-siml} 
\end{equation} 
 
The value $s_y^o$ is used to prune from $P_I$ all those elements whose score $s_y$ is lower than $s_y^o$. The intuitive idea behind this procedure is that we relax the inter-classifier competition, collecting more boxes in those  images associated with multiple labels (there is no competition among  classes contained in $Y$). However, we impose that a classifier for class $y$, $y \in Y$, should be more confident than the strongest {\em noisy} classifier firing on $z_y^o$. 
Note also that we do not need to modify the class-selection strategy because $e(c)$ (Line 6) 
can be computed using Eq.~\ref{class-easiness} which applies also to the case in which a given image $I \in P$ is associated with multiple boxes. 
Intuitively, in {\em SP-SIML}, both the class selection  and intra-image box pruning   are  based on the competition between classifiers in $Y$ and classifiers in $\overline{Y}$.
Apart from Lines 3-5, 
 modified as explained above, {\em SP-SIML} is identical to {\em SP}, including score-based image selection.

{\bf Multiple-label and multiple-instance.} 
{\em SP-MIML} is a Multiple-Instance-Multiple-Label extension of {\em SP-SIML}, where more than one box {\em of the same category} $y \in Y$ can be collected from the same image $I$.  {\em SP-MIML} is obtained applying standard Non-Maxima Suppression (NMS) over $f_{W_{t-1}}(I, B(I))$ and then, for each $y \in Y$, keeping all those boxes whose value is higher than $s_y^o$ (thus, no ad hoc, manually tuned threshold on the box scores after NMS needs to be used). Also in this case, we use both score-based image selection 
(Lines 10-11) 
and class-selection and we modify only Lines 3-5 
 in Alg.~\ref{alg.SelfPaced.training} to obtain {\em SP-MIML}.
 
 We test {\em SP-SIML}  and {\em SP-MIML} on Pascal VOC 2007 because Pascal VOC images contain more objects and more labels on average with respect to ILSVRC 2013. In Tab.~\ref{experiments:siml-miml}
we show the results which have been obtained using $W_0^P$ as the initialization and using the same  hyper-parameter values used in the other  experiments (see Sec.~\ref{WeaklySelf-Paced}).

\begin{table}[h]
\centering
\begin{tabular}{|c|c|}
\hline
    Method     		& mAP 		\\ \hline\hline
  {\em SP-SIML}	  	&  33.80     	\\ \hline
  {\em SP-MIML} 	& 34.43	     	\\ \hline
  {\em SP} 	  		& 	38.11      	\\ \hline
\end{tabular}
\caption{Results (mAP $\%$) on the Pascal VOC 2007  {\em test} 
set using relaxed versions of the inter-classifier competition.}
\label{experiments:siml-miml}
\end{table}

Unexpectedly,  {\em SP} largely outperforms both {\em SP-SIML} and {\em SP-MIML}, despite more data are used in the latter two versions (because more boxes on average are collected from a single image $I$). 
Our interpretation of this result is that the inter-classifier competition is very important for  a prudent choice of the latent boxes, forcing the system to select a box $z_y^I$ only when its classification confidence $s_y^I$ is very high.
To confirm this hypothesis, we computed the average CorLoc  over all the self-paced iterations and using {\em SP}, {\em SP-SIML} and {\em SP-MIML}. Consistently in all $M = 4$ iterations the CorLoc of 
both {\em SP-SIML} and {\em SP-MIML} is much smaller than the corresponding values obtained with 
{\em SP}, which means that the mined pseudo-ground truth is less precise.
 A visual inspection of randomly selected images further confirmed that most of the times the additional boxes selected in both {\em SP-SIML} and {\em SP-MIML} with respect to $z_y^I$ are inaccurately localized or completely wrong. 
However, although the multi-label extensions of {\em SP} underperformed with respect to {\em SP}, their final mAP is comparable with the state of the art on Pascal VOC 07 (see Tab.~\ref{tab.Pascal-mAP}). 


\subsection{Precision of the selected subsets of training data}
\label{Precision}

In this subsection we  evaluate the number of ``correct'' samples selected  for training the network. To this aim we adopt the  evaluation protocol suggested in \cite{DBLP:conf/cvpr/HoffmanPDS15}, where the authors use ILSVRC 2013 {\em val1} and 
a Precision metric. The latter is similar to CorLoc, the difference being that in CorLoc one latent box ($z_y$) is computed for each label $y \in Y$ associated with a training image, while Precision is based on extracting one single latent box ($z_y^I$) per image. Using Precision $@ 0.5$ IoU we can measure the quantity of latent boxes actually used during training which sufficiently overlap with a real ground truth box with the correct class.

In Tab.~\ref{experiments:tab1} we show the results, where {\em Precision} is the percentage of correct image samples over  all the images included in the training set $T_t$.
In case of \cite{DBLP:conf/cvpr/HoffmanPDS15}, Precision is computed with respect to the whole 
{\em val1} because no subset selection is done in that work.
$T_1$ in Tab.~\ref{experiments:tab1} is the dataset obtained with the initialization model $W_0$ 
and used to train $W_1$,
while $T_4$ is the dataset obtained with $W_3$ and used in the last iteration to train $W_4$.   As shown in the table, Precision in $T_4$ is largely improved with respect to Precision in $T_1$.
Precision in $T_4$ is  much higher than the Precision obtained by \cite{DBLP:conf/cvpr/HoffmanPDS15},  even when {\em object-level} annotation for 100 over 200 categories is used as auxiliary data during training.

\begin{figure}
	\centering
	\includegraphics[width =\linewidth]{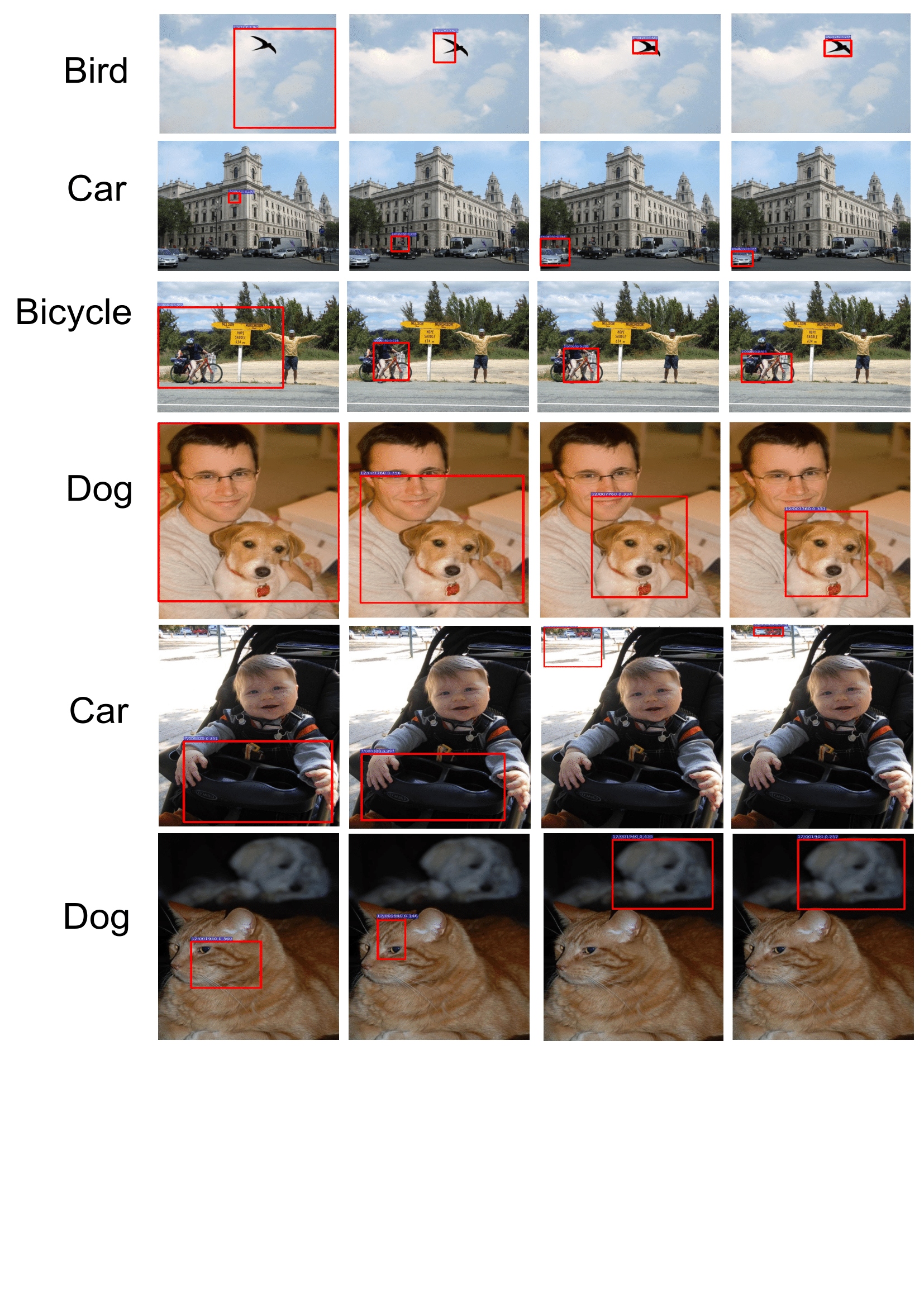}
\vspace{-80pt}
\caption{Qualitative results: visualizations of the  class-specific top-score box $z_y$   in the four  self-paced iterations (chronologically ordered from left to right) with respect to different training images and labels $y$ (leftmost column).}
\label{intro:qualitative_results}
\end{figure}

\begin{figure}
	\centering
	\includegraphics[width =\linewidth]{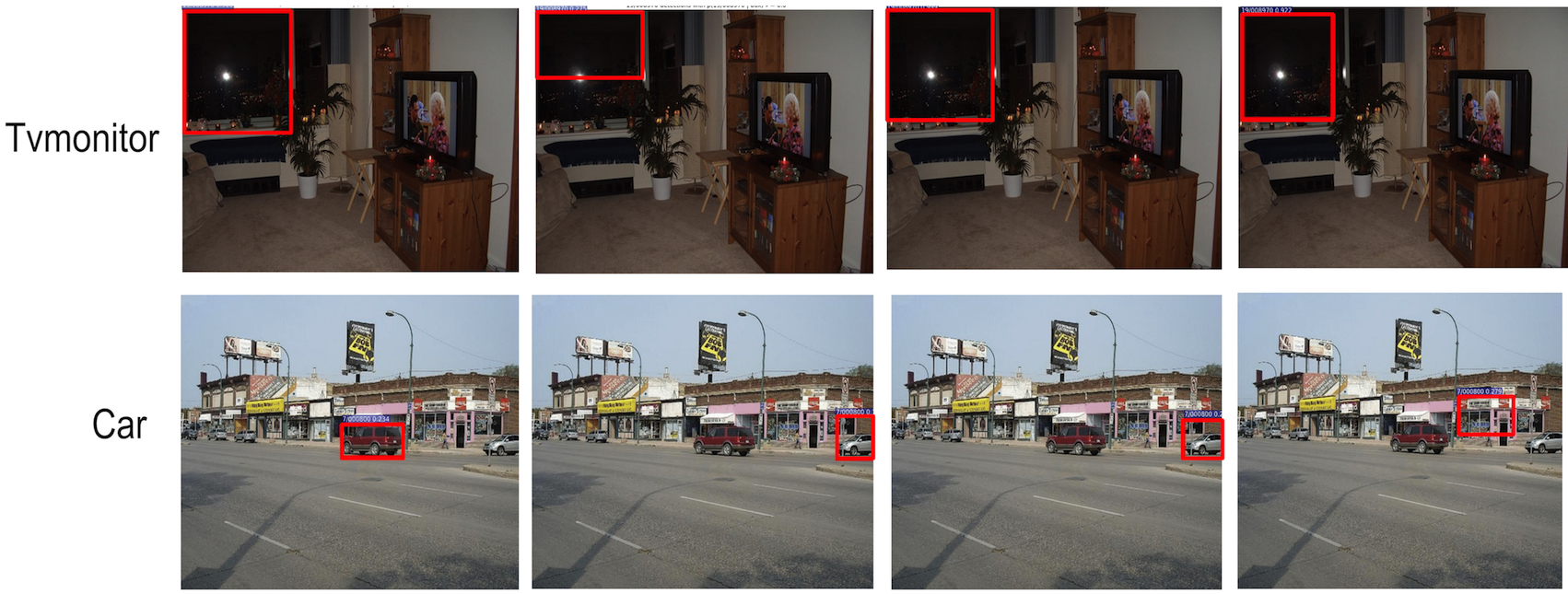}
\caption{Other qualitative results in which the evolution over time of the class-specific top-score box ($z_y$) of the network did {\em not} succeed in localizing the true objects into the images.}
\label{intro:qualitative_resultsNeg}
\end{figure}

In Fig.~\ref{intro:qualitative_results}-\ref{intro:qualitative_resultsNeg} we show 
 the evolution of the class-specific top-score box $z_y$ for the same image $I$  over the four self-paced iterations.
 The new localizations usually improve with respect to the previous ones.

\begin{table*}[!htbp]
\centering
\begin{tabular}{|c|c|}
\hline
       Method   			& Precision	(IoU $> 0.5$) \\ \hline\hline
  Hoffman et al. \cite{DBLP:conf/cvpr/HoffmanPDS15} {\em without} auxiliary strongly supervised data	  			& 	26.10   	\\ \hline
  Hoffman et al. \cite{DBLP:conf/cvpr/HoffmanPDS15} {\em with} auxiliary strongly supervised data	  			& 	28.81   	\\ \hline
  {\em SP} ($T_1$)	  			& 	20.55   	\\ \hline
  {\em SP} ($T_4$)	  			& 	37.01   	\\ \hline
\end{tabular}
\hspace{2cm}
\caption{Precision of the selected boxes used for training. In  {\em SP} the Precision value is computed over the elements in $T_t$, which is a subset of ILSVRC 2013 {\em val1}, while in \cite{DBLP:conf/cvpr/HoffmanPDS15} Precision is computed over the whole {\em val1}. However, the comparison is fair because, differently from \cite{DBLP:conf/cvpr/HoffmanPDS15}, we do {\em not} use the whole {\em val1} for training but only the subset $T_t$, thus the quality of the training boxes should be compared with only those samples actually used for training.}
\label{experiments:tab1}
\end{table*}

\section{Conclusions}
\label{Conclusions}

We proposed a self-paced learning based protocol for deep networks in a WSD scenario, aiming at reducing the amount of noise while training the DN.
Our training protocol 
 extends the self-paced learning paradigm by introducing: (1) inter-classifier competition as a powerful mechanism to reduce noise, (2)
   class-selection, in which the easiest classes are trained first, and (3) the use of the Fast-RCNN regression layer for the implicit modification of the bag of boxes.

While in the past self-paced learning strategies have been successfully adopted for other classifier types (mainly SVMs), we are the first showing that this paradigm can  be successfully utilized also for an end-to-end  training of deep networks.
Despite the reduced sizes of the initial training subsets, typical in a self-paced strategy, 
we have empirically shown that our training protocol not only does not suffer for overfitting, but it benefits from the reduced noise, by largely boosting the final accuracy with respect to different initialization methods.

Using the proposed training protocol
we achieved state-of-the-art results on  common WSD benchmarks:
ILSVRC 2013,
 Pascal VOC 2007 and VOC 2010.

Finally, we presented a detailed analysis of the main components of our proposed training protocol, comparing {\em SP} with both simplified and more sophisticated versions of the same approach, with the goal of showing the importance of our design choices and to allow other authors to build on our method, possibly choosing those components which best fit with other application scenarios.

\section*{Acknowledgements}
We want to thank the NVIDIA Corporation for the  donation  of  the  GPUs  used  in this project.

{\small
\bibliographystyle{ieee}
\bibliography{bibliografia41}

\begin{thebibliography}{10}\itemsep=-1pt

\bibitem{Bazzani:WACV16}
L.~Bazzani, A.~Bergamo, D.~Anguelov, and L.~Torresani.
\newblock Self-taught object localization with deep networks.
\newblock In {\em IEEE Winter Conference on Applications of Computer Vision
  (WACV)}, 2016.

\bibitem{DBLP:conf/icml/BengioLCW09}
Y.~Bengio, J.~Louradour, R.~Collobert, and J.~Weston.
\newblock Curriculum learning.
\newblock In {\em ICML}, pages 41--48, 2009.

\bibitem{Bilen1}
H.~Bilen, M.~Pedersoli, and T.~Tuytelaars.
\newblock Weakly supervised object detection with posterior regularization.
\newblock In {\em BMVC}, 2014.

\bibitem{DBLP:conf/cvpr/BilenPT15}
H.~Bilen, M.~Pedersoli, and T.~Tuytelaars.
\newblock Weakly supervised object detection with convex clustering.
\newblock In {\em CVPR}, 2015.

\bibitem{Bilen16}
H.~Bilen and A.~Vedaldi.
\newblock Weakly supervised deep detection networks.
\newblock In {\em CVPR}, 2016.

\bibitem{DBLP:conf/bmvc/ChatfieldSVZ14}
K.~Chatfield, K.~Simonyan, A.~Vedaldi, and A.~Zisserman.
\newblock Return of the devil in the details: Delving deep into convolutional
  nets.
\newblock In {\em BMVC}, 2014.

\bibitem{DBLP:journals/corr/ChenG15a}
X.~Chen and A.~Gupta.
\newblock Webly supervised learning of convolutional networks.
\newblock In {\em ICCV}, 2015.

\bibitem{jacob}
R.~G. Cinbis, J.~J. Verbeek, and C.~Schmid.
\newblock Multi-fold {MIL} training for weakly supervised object localization.
\newblock In {\em CVPR}, pages 2409--2416, 2014.

\bibitem{DBLP:journals/pami/CinbisVS17}
R.~G. Cinbis, J.~J. Verbeek, and C.~Schmid.
\newblock Weakly supervised object localization with multi-fold multiple
  instance learning.
\newblock {\em {IEEE} Trans. Pattern Anal. Mach. Intell.}, 39(1):189--203,
  2017.

\bibitem{pascal-voc-2007}
M.~Everingham, L.~Van~Gool, C.~K. Williams, J.~Winn, and A.~Zisserman.
\newblock The {Pascal} {V}isual {O}bject {C}lasses {C}hallenge 2007 {(VOC
  2007)} {R}esults.

\bibitem{pascal-voc-2010}
M.~Everingham, L.~Van~Gool, C.~K. Williams, J.~Winn, and A.~Zisserman.
\newblock The {Pascal} {V}isual {O}bject {C}lasses {(VOC)} challenge.
\newblock {\em IJCV}, 88(2):303?--338, 2010.

\bibitem{DBLP:conf/iccv/Girshick15}
R.~B. Girshick.
\newblock Fast {R-CNN}.
\newblock In {\em ICCV}, 2015.

\bibitem{DBLP:conf/cvpr/GirshickDDM14}
R.~B. Girshick, J.~Donahue, T.~Darrell, and J.~Malik.
\newblock Rich feature hierarchies for accurate object detection and semantic
  segmentation.
\newblock In {\em CVPR}, 2014.

\bibitem{DBLP:journals/corr/GkioxariGM15}
G.~Gkioxari, R.~B. Girshick, and J.~Malik.
\newblock Contextual action recognition with {R*CNN}.
\newblock In {\em ICCV}, 2015.

\bibitem{DBLP:conf/iccv/HeGDG17}
K.~He, G.~Gkioxari, P.~Doll{\'{a}}r, and R.~B. Girshick.
\newblock Mask {R-CNN}.
\newblock In {\em ICCV}, 2017.

\bibitem{DBLP:journals/corr/HoffmanGTDGDS14}
J.~Hoffman, S.~Guadarrama, E.~Tzeng, R.~Hu, J.~Donahue, R.~B. Girshick,
  T.~Darrell, and K.~Saenko.
\newblock {LSDA:} large scale detection through adaptation.
\newblock In {\em NIPS}, 2014.

\bibitem{DBLP:conf/cvpr/HoffmanPDS15}
J.~Hoffman, D.~Pathak, T.~Darrell, and K.~Saenko.
\newblock Detector discovery in the wild: Joint multiple instance and
  representation learning.
\newblock In {\em CVPR}, pages 2883--2891, 2015.

\bibitem{DBLP:conf/nips/JiangMYLSH14}
L.~Jiang, D.~Meng, S.~Yu, Z.~Lan, S.~Shan, and A.~G. Hauptmann.
\newblock Self-paced learning with diversity.
\newblock In {\em NIPS}, pages 2078--2086, 2014.

\bibitem{kantorov2016}
V.~Kantorov, M.~Oquab, M.~Cho, and I.~Laptev.
\newblock {ContextLocNet:} context-aware deep network models for weakly
  supervised localization.
\newblock In {\em ECCV}, 2016.

\bibitem{ECCV12_Khosla}
A.~Khosla, T.~Zhou, T.~Malisiewicz, A.~Efros, and A.~Torralba.
\newblock Undoing the damage of dataset bias.
\newblock In {\em ECCV}, 2012.

\bibitem{DBLP:conf/nips/KrizhevskySH12}
A.~Krizhevsky, I.~Sutskever, and G.~E. Hinton.
\newblock {ImageNet} classification with deep convolutional neural networks.
\newblock In {\em NIPS}, 2012.

\bibitem{DBLP:conf/nips/KumarPK10}
M.~P. Kumar, B.~Packer, and D.~Koller.
\newblock Self-paced learning for latent variable models.
\newblock In {\em NIPS}, pages 1189--1197, 2010.

\bibitem{DBLP:journals/corr/LapedrizaPBT13}
{\`{A}}.~Lapedriza, H.~Pirsiavash, Z.~Bylinskii, and A.~Torralba.
\newblock Are all training examples equally valuable?
\newblock {\em arxiv:1311.6510}, 2013.

\bibitem{DBLP:conf/cvpr/LeeG11}
Y.~J. Lee and K.~Grauman.
\newblock Learning the easy things first: Self-paced visual category discovery.
\newblock In {\em CVPR}, pages 1721--1728, 2011.

\bibitem{Li_2016_CVPR}
D.~Li, J.-B. Huang, Y.~Li, S.~Wang, and M.-H. Yang.
\newblock Weakly supervised object localization with progressive domain
  adaptation.
\newblock In {\em CVPR}, 2016.

\bibitem{DBLP:journals/corr/LiangLWLLY14}
X.~Liang, S.~Liu, Y.~Wei, L.~Liu, L.~Lin, and S.~Yan.
\newblock Towards computational baby learning: {A} weakly-supervised approach
  for object detection.
\newblock In {\em ICCV}, 2015.

\bibitem{malisiewicz-iccv11}
T.~Malisiewicz, A.~Gupta, and A.~A. Efros.
\newblock Ensemble of {Exemplar-SVMs} for object detection and beyond.
\newblock In {\em ICCV}, 2011.

\bibitem{DBLP:journals/pr/NguyenTTR14}
M.~H. Nguyen, L.~Torresani, F.~D. la~Torre, and C.~Rother.
\newblock Learning discriminative localization from weakly labeled data.
\newblock {\em Pattern Recognition}, 47(3):1523--1534, 2014.

\bibitem{DBLP:conf/cvpr/OquabBLS15}
M.~Oquab, L.~Bottou, I.~Laptev, and J.~Sivic.
\newblock Is object localization for free? - {W}eakly-supervised learning with
  convolutional neural networks.
\newblock In {\em CVPR}, pages 685--694, 2015.

\bibitem{DBLP:conf/cvpr/PentinaSL15}
A.~Pentina, V.~Sharmanska, and C.~H. Lampert.
\newblock Curriculum learning of multiple tasks.
\newblock In {\em CVPR}, pages 5492--5500, 2015.

\bibitem{DBLP:journals/corr/RussakovskyDSKSMHKKBBF14}
O.~Russakovsky, J.~Deng, H.~Su, J.~Krause, S.~Satheesh, S.~Ma, Z.~Huang,
  A.~Karpathy, A.~Khosla, M.~S. Bernstein, A.~C. Berg, and F.~Li.
\newblock Imagenet large scale visual recognition challenge.
\newblock {\em arxiv:1409.0575}, 2014.

\bibitem{DBLP:conf/eccv/Sangineto14}
E.~Sangineto.
\newblock Statistical and spatial consensus collection for detector adaptation.
\newblock In {\em ECCV}, pages 456--471, 2014.

\bibitem{DBLP:conf/cvpr/SchroffKP15}
F.~Schroff, D.~Kalenichenko, and J.~Philbin.
\newblock {FaceNet: A} unified embedding for face recognition and clustering.
\newblock In {\em CVPR}, pages 815--823, 2015.

\bibitem{DBLP:conf/eccv/ShiF16}
M.~Shi and V.~Ferrari.
\newblock Weakly supervised object localization using size estimates.
\newblock In {\em ECCV}, pages 105--121, 2016.

\bibitem{Shrivastava_2016_CVPR}
A.~Shrivastava, A.~Gupta, and R.~Girshick.
\newblock Training region-based object detectors with online hard example
  mining.
\newblock In {\em CVPR}, 2016.

\bibitem{DBLP:journals/corr/SimonyanZ14a}
K.~Simonyan and A.~Zisserman.
\newblock Very deep convolutional networks for large-scale image recognition.
\newblock {\em arXiv:1409.1556}, 2014.

\bibitem{Song1}
H.~O. Song, R.~Girshick, S.~Jegelka, J.~Mairal, Z.~Harchaoui, and T.~Darrell.
\newblock On learning to localize objects with minimal supervision.
\newblock In {\em ICML}, 2014.

\bibitem{Song2}
H.~O. Song, Y.~J. Lee, S.~Jegelka, and T.~Darrell.
\newblock Weakly-supervised discovery of visual pattern configurations.
\newblock In {\em NIPS}, 2014.

\bibitem{DBLP:conf/cvpr/SupancicR13}
J.~S. {Supancic III} and D.~Ramanan.
\newblock Self-paced learning for long-term tracking.
\newblock In {\em CVPR}, pages 2379--2386, 2013.

\bibitem{attentionBMVC2016}
E.~W. Teh, M.~Rochan, and Y.~Wang.
\newblock Attention networks for weakly supervised object localization.
\newblock In {\em BMVC}, 2016.

\bibitem{DBLP:journals/ijcv/UijlingsSGS13}
J.~R.~R. Uijlings, K.~E.~A. van~de Sande, T.~Gevers, and A.~W.~M. Smeulders.
\newblock Selective search for object recognition.
\newblock {\em International Journal of Computer Vision}, 104(2):154--171,
  2013.

\bibitem{DBLP:journals/tip/WangHRZM15}
C.~Wang, K.~Huang, W.~Ren, J.~Zhang, and S.~J. Maybank.
\newblock Large-scale weakly supervised object localization via latent category
  learning.
\newblock {\em {IEEE} Transactions on Image Processing}, 24(4):1371--1385,
  2015.

\bibitem{DBLP:journals/corr/WeiLCSCZY15}
Y.~Wei, X.~Liang, Y.~Chen, X.~Shen, M.~Cheng, J.~Feng, Y.~Zhao, and S.~Yan.
\newblock {STC:} {A} simple to complex framework for weakly-supervised semantic
  segmentation.
\newblock {\em {IEEE} Trans. Pattern Anal. Mach. Intell.}, 39(11):2314--2320,
  2017.

\bibitem{DBLP:journals/corr/ZarembaS14}
W.~Zaremba and I.~Sutskever.
\newblock Learning to execute.
\newblock {\em arXiv:1410.4615}, 2014.

\bibitem{DBLP:conf/ijcai/ZhangMZH16}
D.~Zhang, D.~Meng, L.~Zhao, and J.~Han.
\newblock Bridging saliency detection to weakly supervised object detection
  based on self-paced curriculum learning.
\newblock In {\em IJCAI}, pages 3538--3544, 2016.

\end{thebibliography}
}


\begin{IEEEbiography}[{\includegraphics[width=1in,height=1.25in,clip,keepaspectratio]{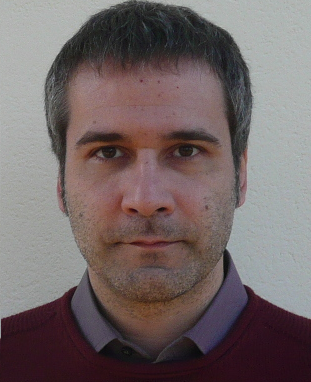}}]{Enver Sangineto}
is a  research fellow  at University of Trento, Department of Information Engineering and Computer Science. 
He received 
his PhD in Computer Engineering from the University of Rome ``La Sapienza''.
After that he has been a post-doctoral researcher at ``La Sapienza'' and at the Italian Institute of Technology in Genova.
 His research interests include object detection and 
learning with minimal human supervision.
\end{IEEEbiography}

\begin{IEEEbiography}[{\includegraphics[width=1in,height=1.25in,clip,keepaspectratio]{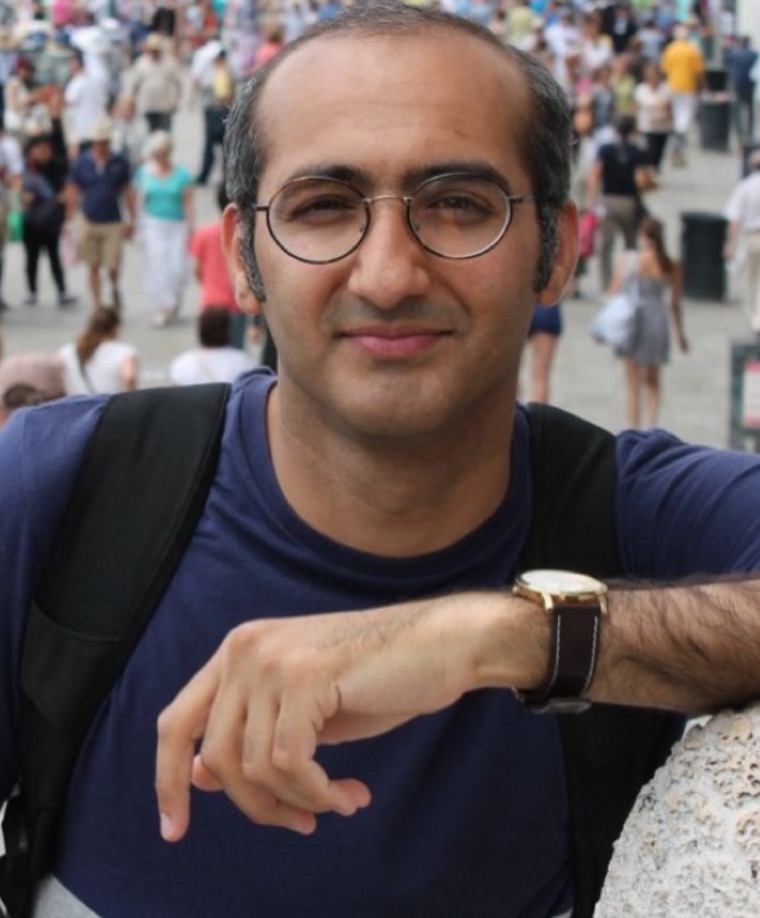}}]{Moin Nabi}
 is a Senior Research Scientist at SAP Machine Learning Research in Berlin. Before that, he was a postdoctoral research fellow in the Deep Relational Learning group at the University of Trento and a visiting researcher in the GRAIL lab at the University of Washington. He holds a PhD from the Italian Institute of Technology. His research lies at the intersection of machine learning, computer vision, and natural language processing with an emphasis on learning Deep Neural Networks with minimal supervision and/or limited data.
\end{IEEEbiography}

\begin{IEEEbiography}[{\includegraphics[width=1in,height=1.25in,clip,keepaspectratio]{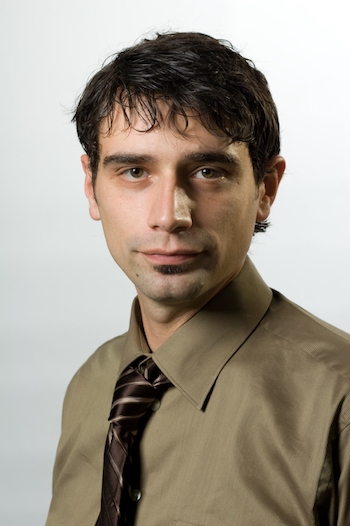}}]{Dubravko Culibrk}
 is a Professor at the University of Novi Sad, Serbia and the director of the Industry/University Center for Collaborative Research of the Faculty of Technical Sciences. His interests are in the domain of computer vision, machine learning and knowledge discovery.
Prof. Culibrk received his B.Eng. degree in Electrical Engineering and an M.Sc. degree in Computer Engineering from the University of Novi Sad, Serbia, in 2000 and 2003, respectively, as well as a Ph.D. degree in computer engineering from Florida Atlantic University, Boca Raton, USA, in 2006. 
\end{IEEEbiography}

\begin{IEEEbiography}[{\includegraphics[width=1in,height=1.25in,clip,keepaspectratio]{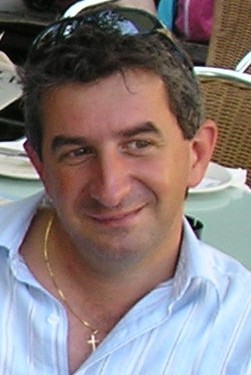}}]{Nicu Sebe}  is Professor with the University of
Trento, Italy, leading the research in the areas of
multimedia information retrieval and human behavior
understanding. 
He was the General Chair of the IEEE FG Conference 2008, ACM Multimedia 2013 and ICMR 2017, and the Program Chair
of the International Conference on Image and Video Retrieval in 2007 and 2010, ACM Multimedia 2007 and 2011, ECCV 2016 and ICCV 2017. He is the Program Chair of ICPR 2020. 
He is a fellow
of the International Association for Pattern Recognition.
\end{IEEEbiography}


\clearpage
\appendix
\section{Appendix}
\label{Appendix}

In Tab.~\ref{table_full} we show the per-category AP obtained by our method on ILSVRC 2013 {\em val2}. Similarly to Sec.~7.1, we analyse 5 different nets ($W_0, ... W_4$), corresponding to the 4 self-paced iterations of Alg.~1 plus the initialization model ($W_0 = W_0^I$, i.e., $Init$), used as a comparison. This is done in order to show that most of the categories progressively improve during training and that this improvement is generalized and not dominated by a few categories. In fact, the AP of 147 out of 200 categories increases when evaluated using $f_{W_4}$ with 
respect to the evaluation obtained using the initial model $f_{W_0}$ (see Sec.~7.1).

This is not a trivial result, because
the combination of class-selection and inter-classifier competition
in {\em SP}
could potentially lead the sample selection process to be dominated by a few strong 
categories in $f_{W_0}$.
For instance, categories like antelope or fox, which have AP 22.09 and 22.61 using  $f_{W_0}$, respectively,  are already strong in the beginning of the learning process and could dominate the selection of new samples in $T_1, T_2, ...$. Conversely, initially very weak classifiers like cream or oboe (AP 3.06 and 1.51 using  $f_{W_0}$, respectively) could be penalized because not able to win the inter-classifier competition or 
because excluded by the class-selection process.
 However, our empirical results show that this harmful domination of the initial strong classifiers does {\em not} happen and that learning is spread over most of  the categories. We believe that this is due to the fact that good classifiers (e.g., antelope) do {\em not} have high scores in an image showing an oboe, because most of the BBs of the oboe's image (background boxes included) have an appearance different from an antelope. Thus even a weak classifier can win the competition on its own samples and ''gain'' new samples to add to the next training set $T_t$ which will finally lead to the improvement the weak classifier.

\clearpage
\begin{table}[]
\caption{Per-class AP on the ILSVRC13 detection {\em val2}.}
\centering
\label{table_full}
\begin{tabular}{l|p{3.5cm}p{1cm}p{1cm}p{1cm}p{1cm}c}
\hline
\#  & \textbf{Category}            & \textbf{W0}    & \textbf{W1}    & \textbf{W2}    & \textbf{W3}    & \textbf{W4}    \\ \hline\hline
1   & accordion             & 31.83 & 31.50 & 32.31 & 27.45 & 33.25 \\
2   & airplane              & 34.55 & 32.93 & 35.55 & 43.35 & 42.61 \\
3   & ant                   & 17.79 & 15.58 & 16.75 & 15.77 & 16.14 \\
4   & antelope              & 22.09 & 24.85 & 30.44 & 32.61 & 31.36 \\
5   & apple                 & 6.07  & 10.32 & 10.54 & 10.96 & 11.48 \\
6   & armadillo             & 32.39 & 32.16 & 33.95 & 37.25 & 37.49 \\
7   & artichoke             & 14.85 & 15.03 & 17.03 & 18.28 & 18.22 \\
8   & axe                   & 0.80  & 0.10  & 1.15  & 2.47  & 1.47  \\
9   & baby bed              & 4.65  & 17.66 & 19.40 & 18.44 & 14.33 \\
10  & backpack              & 1.39  & 1.69  & 1.90  & 2.26  & 2.06  \\
11  & bagel                 & 6.11  & 10.32 & 11.64 & 11.57 & 11.81 \\
12  & balance beam          & 0.03  & 0.12  & 0.15  & 0.20  & 0.19  \\
13  & banana                & 7.13  & 6.78  & 8.73  & 8.82  & 13.67 \\
14  & band aid              & 0.27  & 3.52  & 0.41  & 0.48  & 0.62  \\
15  & banjo                 & 18.92 & 11.32 & 11.18 & 6.22  & 6.29  \\
16  & baseball              & 7.21  & 18.38 & 23.41 & 27.91 & 27.90 \\
17  & basketball            & 0.00  & 0.01  & 0.00  & 0.00  & 0.00  \\
18  & bathing cap           & 2.38  & 4.89  & 2.44  & 1.40  & 1.48  \\
19  & beaker                & 4.17  & 10.02 & 10.83 & 11.70 & 11.20 \\
20  & bear                  & 31.87 & 29.54 & 31.77 & 28.81 & 29.42 \\
21  & bee                   & 12.85 & 15.61 & 15.85 & 14.04 & 19.51 \\
22  & bell pepper           & 9.21  & 11.48 & 7.92  & 8.84  & 8.71  \\
23  & bench                 & 1.82  & 3.26  & 1.07  & 0.52  & 0.59  \\
24  & bicycle               & 19.05 & 21.71 & 21.87 & 22.05 & 21.43 \\
25  & binder                & 3.83  & 3.31  & 4.28  & 4.38  & 4.90  \\
26  & bird                  & 39.59 & 43.81 & 47.81 & 47.24 & 46.97 \\
27  & bookshelf             & 4.92  & 7.76  & 9.55  & 7.62  & 7.68  \\
28  & bow                   & 0.89  & 1.10  & 1.79  & 1.24  & 1.20  \\
29  & bow tie               & 0.23  & 0.12  & 0.10  & 1.88  & 1.49  \\
30  & bowl                  & 5.08  & 5.37  & 6.29  & 8.40  & 7.58  \\
31  & brassiere             & 9.01  & 10.68 & 11.26 & 11.18 & 11.67 \\
32  & burrito               & 3.03  & 2.16  & 5.01  & 5.00  & 5.33  \\
33  & bus                   & 28.70 & 37.36 & 36.12 & 36.26 & 34.82 \\
34  & butterfly             & 60.85 & 60.18 & 67.47 & 67.87 & 67.47 \\
35  & camel                 & 6.66  & 7.92  & 9.59  & 10.65 & 10.33 \\
36  & can opener            & 8.48  & 11.41 & 12.15 & 10.64 & 12.79 \\
37  & car                   & 21.11 & 22.33 & 22.98 & 24.82 & 25.07 \\
38  & cart                  & 12.62 & 11.46 & 11.52 & 11.68 & 11.01 \\
39  & cattle                & 9.97  & 3.35  & 6.24  & 5.39  & 6.27  \\
40  & cello                 & 6.49  & 12.75 & 13.70 & 16.99 & 16.92 \\
41  & centipede             & 13.34 & 12.75 & 16.59 & 16.26 & 19.86 \\
42  & chain saw             & 0.09  & 0.70  & 2.08  & 1.80  & 2.17  \\
43  & chair                 & 4.03  & 5.13  & 6.10  & 5.60  & 5.53  \\
44  & chime                 & 6.94  & 10.43 & 7.81  & 3.31  & 6.34  \\
45  & cocktail shaker       & 11.75 & 10.23 & 13.56 & 14.93 & 15.37 \\
46  & coffee maker          & 3.31  & 12.70 & 12.36 & 15.27 & 11.37 \\
47  & computer keyboard     & 3.18  & 5.89  & 9.20  & 11.36 & 11.78 \\
48  & computer mouse        & 1.70  & 1.71  & 1.71  & 2.78  & 2.21  \\
49  & corkscrew             & 13.03 & 9.63  & 10.76 & 13.17 & 13.06 \\
50  & cream                 & 3.06  & 5.61  & 6.80  & 10.82 & 10.65 \\
\end{tabular}
\end{table}
\clearpage
\begin{table}[]
\centering
\begin{tabular}{l|p{3.5cm}p{1cm}p{1cm}p{1cm}p{1cm}c}
\hline
\#  & \textbf{Category}            & \textbf{W0}    & \textbf{W1}    & \textbf{W2}    & \textbf{W3}    & \textbf{W4}    \\ \hline\hline
51  & croquet ball          & 1.72  & 0.66  & 0.28  & 0.16  & 0.25  \\
52  & crutch                & 1.34  & 1.96  & 1.79  & 1.85  & 2.78  \\
53  & cucumber              & 5.22  & 4.52  & 3.59  & 3.65  & 3.41  \\
54  & cup or mug            & 7.51  & 7.28  & 11.99 & 12.10 & 12.67 \\
55  & diaper                & 1.46  & 4.60  & 3.13  & 3.43  & 2.96  \\
56  & digital clock         & 4.62  & 13.05 & 20.51 & 19.04 & 19.35 \\
57  & dishwasher            & 7.79  & 0.79  & 0.96  & 0.70  & 0.67  \\
58  & dog                   & 16.80 & 15.59 & 17.34 & 19.32 & 17.35 \\
59  & domestic cat          & 4.10  & 3.87  & 3.15  & 4.87  & 3.23  \\
60  & dragonfly             & 18.44 & 16.02 & 26.51 & 25.22 & 25.29 \\
61  & drum                  & 0.23  & 0.42  & 0.57  & 0.56  & 0.57  \\
62  & dumbbell              & 2.56  & 1.49  & 1.43  & 1.32  & 1.36  \\
63  & electric fan          & 28.65 & 16.71 & 16.83 & 20.57 & 21.10 \\
64  & elephant              & 31.17 & 38.46 & 34.92 & 37.21 & 34.44 \\
65  & face powder           & 9.96  & 7.85  & 6.27  & 5.19  & 4.65  \\
66  & fig                   & 7.56  & 7.69  & 6.70  & 4.59  & 4.10  \\
67  & filing cabinet        & 3.60  & 6.26  & 7.72  & 6.92  & 7.74  \\
68  & flower pot            & 0.24  & 0.39  & 0.28  & 0.27  & 0.27  \\
69  & flute                 & 0.08  & 1.68  & 1.67  & 1.60  & 1.57  \\
70  & fox                   & 22.61 & 28.19 & 28.73 & 32.96 & 36.23 \\
71  & french horn           & 9.17  & 10.71 & 10.50 & 16.79 & 12.28 \\
72  & frog                  & 22.53 & 28.31 & 29.88 & 29.37 & 30.08 \\
73  & frying pan            & 5.08  & 4.03  & 4.41  & 5.79  & 5.70  \\
74  & giant panda           & 34.60 & 46.05 & 46.31 & 26.62 & 26.71 \\
75  & goldfish              & 5.37  & 9.75  & 7.69  & 9.40  & 9.42  \\
76  & golf ball             & 17.15 & 22.84 & 23.18 & 23.10 & 22.91 \\
77  & golfcart              & 32.61 & 24.31 & 28.82 & 31.24 & 33.59 \\
78  & guacamole             & 13.27 & 8.80  & 8.33  & 8.65  & 8.28  \\
79  & guitar                & 5.49  & 5.35  & 6.69  & 4.91  & 4.82  \\
80  & hair dryer            & 3.74  & 0.44  & 0.86  & 1.36  & 1.28  \\
81  & hair spray            & 1.71  & 2.57  & 3.85  & 3.75  & 4.94  \\
82  & hamburger             & 15.57 & 31.05 & 25.96 & 17.97 & 17.92 \\
83  & hammer                & 0.72  & 0.54  & 0.64  & 1.27  & 1.58  \\
84  & hamster               & 41.87 & 31.78 & 42.78 & 39.35 & 39.29 \\
85  & harmonica             & 0.49  & 0.94  & 0.76  & 0.62  & 0.50  \\
86  & harp                  & 21.65 & 30.85 & 34.71 & 38.22 & 37.95 \\
87  & hat with a wide brim  & 6.10  & 10.00 & 11.51 & 9.85  & 11.37 \\
88  & head cabbage          & 4.92  & 8.11  & 8.76  & 9.63  & 9.64  \\
89  & helmet                & 5.08  & 4.10  & 4.51  & 5.16  & 5.43  \\
90  & hippopotamus          & 21.24 & 24.43 & 30.54 & 27.03 & 27.58 \\
91  & horizontal bar        & 0.04  & 0.07  & 0.07  & 0.05  & 0.05  \\
92  & horse                 & 7.34  & 6.74  & 9.30  & 9.06  & 8.90  \\
93  & hotdog                & 4.06  & 4.74  & 4.12  & 4.13  & 3.77  \\
94  & iPod                  & 15.80 & 26.91 & 26.39 & 25.08 & 25.70 \\
95  & isopod                & 10.06 & 13.45 & 12.88 & 13.31 & 10.11 \\
96  & jellyfish             & 7.32  & 3.41  & 2.76  & 4.39  & 4.36  \\
97  & koala bear            & 37.38 & 46.53 & 50.72 & 56.96 & 52.95 \\
98  & ladle                 & 0.47  & 1.75  & 0.52  & 0.57  & 0.58  \\
99  & ladybug               & 11.74 & 13.94 & 13.06 & 13.69 & 13.35 \\
100 & lamp                  & 0.99  & 0.87  & 0.61  & 0.53  & 0.58  \\
\end{tabular}
\end{table}
\clearpage
\begin{table}[]
\centering
\begin{tabular}{l|p{3.5cm}p{1cm}p{1cm}p{1cm}p{1cm}c}
\hline
\#  & \textbf{Category}            & \textbf{W0}    & \textbf{W1}    & \textbf{W2}    & \textbf{W3}    & \textbf{W4}    \\ \hline\hline
101 & laptop                & 7.60  & 14.38 & 15.47 & 18.66 & 18.75 \\
102 & lemon                 & 5.81  & 10.57 & 14.28 & 14.60 & 15.03 \\
103 & lion                  & 4.51  & 0.80  & 2.56  & 2.04  & 1.90  \\
104 & lipstick              & 3.50  & 3.74  & 3.14  & 4.04  & 3.85  \\
105 & lizard                & 6.39  & 11.37 & 13.14 & 13.30 & 13.78 \\
106 & lobster               & 5.12  & 8.48  & 17.80 & 19.80 & 15.46 \\
107 & maillot               & 0.20  & 1.64  & 1.35  & 1.05  & 1.36  \\
108 & maraca                & 3.36  & 2.65  & 1.53  & 1.57  & 1.55  \\
109 & microphone            & 0.07  & 0.03  & 0.06  & 0.10  & 0.11  \\
110 & microwave             & 19.30 & 18.18 & 14.61 & 14.25 & 14.32 \\
111 & milk can              & 16.62 & 19.41 & 21.39 & 14.75 & 17.97 \\
112 & miniskirt             & 0.74  & 0.89  & 0.48  & 0.19  & 0.20  \\
113 & monkey                & 17.49 & 18.81 & 24.26 & 26.50 & 24.46 \\
114 & motorcycle            & 19.21 & 19.08 & 26.79 & 26.96 & 26.70 \\
115 & mushroom              & 13.53 & 16.11 & 17.66 & 18.54 & 18.66 \\
116 & nail                  & 0.07  & 0.44  & 0.21  & 0.13  & 0.19  \\
117 & neck brace            & 0.11  & 6.26  & 1.29  & 0.21  & 0.20  \\
118 & oboe                  & 1.51  & 7.16  & 7.52  & 7.47  & 7.51  \\
119 & orange                & 3.85  & 1.69  & 5.58  & 6.42  & 6.14  \\
120 & otter                 & 2.73  & 2.19  & 5.89  & 8.51  & 9.02  \\
121 & pencil box            & 2.89  & 6.76  & 6.25  & 4.28  & 4.84  \\
122 & pencil sharpener      & 1.49  & 2.35  & 2.05  & 1.42  & 1.59  \\
123 & perfume               & 11.55 & 2.33  & 5.54  & 5.09  & 5.48  \\
124 & person                & 0.14  & 0.27  & 0.37  & 0.47  & 0.47  \\
125 & piano                 & 5.06  & 3.21  & 9.26  & 11.43 & 11.45 \\
126 & pineapple             & 6.20  & 12.69 & 12.49 & 11.82 & 11.92 \\
127 & ping-pong ball        & 0.01  & 0.01  & 0.01  & 0.01  & 0.01  \\
128 & pitcher               & 3.66  & 5.98  & 7.53  & 8.35  & 8.08  \\
129 & pizza                 & 8.89  & 17.77 & 14.55 & 11.71 & 11.55 \\
130 & plastic bag           & 0.39  & 2.21  & 2.74  & 2.82  & 2.55  \\
131 & plate rack            & 1.78  & 1.08  & 2.62  & 6.29  & 6.19  \\
132 & pomegranate           & 8.72  & 8.89  & 9.15  & 9.24  & 9.42  \\
133 & popsicle              & 0.06  & 0.04  & 0.01  & 0.01  & 0.01  \\
134 & porcupine             & 24.08 & 26.20 & 33.64 & 29.51 & 31.30 \\
135 & power drill           & 0.83  & 1.01  & 5.40  & 7.14  & 7.10  \\
136 & pretzel               & 3.33  & 4.79  & 6.04  & 6.20  & 6.40  \\
137 & printer               & 6.17  & 2.79  & 2.37  & 2.42  & 2.29  \\
138 & puck                  & 0.01  & 0.01  & 0.01  & 0.01  & 0.01  \\
139 & punching bag          & 0.58  & 2.33  & 3.65  & 4.35  & 4.03  \\
140 & purse                 & 0.82  & 1.23  & 1.32  & 1.28  & 1.29  \\
141 & rabbit                & 43.34 & 48.27 & 47.09 & 48.31 & 48.41 \\
142 & racket                & 0.05  & 0.06  & 0.08  & 0.08  & 0.06  \\
143 & ray                   & 10.26 & 14.08 & 19.76 & 20.22 & 22.67 \\
144 & red panda             & 16.13 & 27.36 & 22.34 & 25.36 & 25.28 \\
145 & refrigerator          & 9.90  & 10.05 & 8.14  & 9.21  & 8.13  \\
146 & remote control        & 17.05 & 26.71 & 32.35 & 24.36 & 25.56 \\
147 & rubber eraser         & 0.01  & 0.01  & 0.01  & 0.01  & 0.01  \\
148 & rugby ball            & 0.04  & 0.06  & 0.05  & 0.04  & 0.04  \\
149 & ruler                 & 0.30  & 1.89  & 3.84  & 3.66  & 3.90  \\
150 & salt or pepper shaker & 9.99  & 8.50  & 7.69  & 5.45  & 6.92  \\
\end{tabular}
\end{table}
\clearpage
\begin{table}[]
\centering
\label{my-label}
\begin{tabular}{l|p{3.5cm}p{1cm}p{1cm}p{1cm}p{1cm}c}
\hline
\#  & \textbf{Category}            & \textbf{W0}    & \textbf{W1}    & \textbf{W2}    & \textbf{W3}    & \textbf{W4}    \\ \hline\hline
151 & saxophone             & 15.90 & 14.08 & 13.23 & 11.66 & 12.78 \\
152 & scorpion              & 15.23 & 17.18 & 22.04 & 18.47 & 16.83 \\
153 & screwdriver           & 0.11  & 0.17  & 0.13  & 0.68  & 0.35  \\
154 & seal                  & 2.22  & 3.18  & 3.33  & 3.30  & 3.96  \\
155 & sheep                 & 18.81 & 14.57 & 18.88 & 19.25 & 18.59 \\
156 & ski                   & 0.09  & 0.02  & 0.14  & 0.06  & 0.06  \\
157 & skunk                 & 6.61  & 9.70  & 9.19  & 12.30 & 10.85 \\
158 & snail                 & 24.51 & 11.68 & 16.11 & 19.22 & 19.46 \\
159 & snake                 & 1.58  & 5.20  & 12.69 & 15.81 & 16.05 \\
160 & snowmobile            & 21.22 & 31.10 & 27.13 & 25.74 & 28.05 \\
161 & snowplow              & 33.11 & 38.49 & 37.20 & 41.74 & 41.12 \\
162 & soap dispenser        & 0.01  & 0.50  & 0.19  & 0.20  & 0.17  \\
163 & soccer ball           & 20.17 & 19.04 & 17.84 & 18.06 & 18.04 \\
164 & sofa                  & 7.65  & 8.22  & 7.14  & 7.66  & 7.71  \\
165 & spatula               & 0.04  & 0.04  & 0.07  & 0.11  & 0.12  \\
166 & squirrel              & 17.91 & 10.72 & 16.12 & 25.52 & 23.18 \\
167 & starfish              & 3.88  & 9.89  & 13.17 & 17.33 & 17.25 \\
168 & stethoscope           & 2.12  & 2.97  & 7.39  & 7.67  & 7.74  \\
169 & stove                 & 0.58  & 1.32  & 1.81  & 2.18  & 2.04  \\
170 & strainer              & 0.14  & 0.26  & 0.87  & 3.64  & 3.76  \\
171 & strawberry            & 4.90  & 7.82  & 7.52  & 7.33  & 7.20  \\
172 & stretcher             & 0.10  & 0.01  & 0.15  & 0.10  & 0.10  \\
173 & sunglasses            & 1.51  & 2.52  & 1.39  & 1.35  & 1.74  \\
174 & swimming trunks       & 0.11  & 0.03  & 0.00  & 0.00  & 0.00  \\
175 & swine                 & 18.70 & 23.02 & 28.07 & 34.95 & 31.21 \\
176 & syringe               & 1.55  & 2.62  & 2.64  & 2.64  & 2.62  \\
177 & table                 & 1.36  & 3.18  & 3.77  & 3.25  & 3.34  \\
178 & tape player           & 11.73 & 8.12  & 8.94  & 9.92  & 10.32 \\
179 & tennis ball           & 8.88  & 7.11  & 1.61  & 1.75  & 1.35  \\
180 & tick                  & 19.72 & 18.95 & 24.49 & 17.99 & 18.74 \\
181 & tie                   & 3.13  & 2.91  & 3.43  & 2.25  & 3.28  \\
182 & tiger                 & 10.38 & 15.88 & 18.89 & 20.83 & 26.02 \\
183 & toaster               & 22.46 & 22.21 & 21.94 & 21.79 & 22.69 \\
184 & traffic light         & 1.20  & 1.14  & 1.03  & 0.82  & 0.90  \\
185 & train                 & 10.28 & 22.47 & 19.73 & 17.29 & 17.68 \\
186 & trombone              & 3.49  & 4.04  & 1.15  & 1.72  & 1.66  \\
187 & trumpet               & 2.66  & 4.12  & 3.75  & 3.83  & 3.88  \\
188 & turtle                & 22.47 & 27.24 & 31.95 & 31.68 & 32.04 \\
189 & tv or monitor         & 20.12 & 29.23 & 33.18 & 33.81 & 33.84 \\
190 & unicycle              & 0.96  & 1.51  & 0.70  & 0.54  & 0.56  \\
191 & vacuum                & 1.49  & 0.50  & 0.48  & 0.67  & 0.37  \\
192 & violin                & 2.02  & 2.06  & 3.19  & 2.42  & 5.22  \\
193 & volleyball            & 0.02  & 0.09  & 0.02  & 0.02  & 0.02  \\
194 & waffle iron           & 2.71  & 2.03  & 2.29  & 2.98  & 4.09  \\
195 & washer                & 41.19 & 35.88 & 34.47 & 30.45 & 31.32 \\
196 & water bottle          & 5.28  & 7.32  & 4.89  & 6.23  & 6.05  \\
197 & watercraft            & 6.62  & 6.55  & 5.68  & 3.65  & 4.39  \\
198 & whale                 & 23.15 & 27.97 & 32.27 & 33.57 & 37.38 \\
199 & wine bottle           & 2.74  & 1.75  & 1.87  & 1.34  & 1.74  \\
200 & zebra                 & 31.42 & 42.00 & 43.26 & 40.14 & 42.17 \\ \hline
 --- & \textbf{mAP}        & \textbf{9.54} & \textbf{10.88} & \textbf{11.87} & \textbf{12.01} & \textbf{12.13} 
\end{tabular}
\end{table}

\end{document}